\documentclass[sigconf]{acmart}

\usepackage{amsmath,graphicx}
\usepackage{multirow}
\usepackage{epstopdf}
\usepackage{rotating}
\usepackage{mathrsfs}

\usepackage{amssymb}

\usepackage{algorithm}
\usepackage{algorithmic}
\usepackage{xcolor}
\usepackage{makecell}
\usepackage{subfigure}
\usepackage{overpic}
\usepackage{array}
\usepackage{hyperref}
\usepackage{balance}
\AtBeginDocument{%
  \providecommand\BibTeX{{%
    \normalfont B\kern-0.5em{\scshape i\kern-0.25em b}\kern-0.8em\TeX}}}

\copyrightyear{2022}
\acmYear{2022}
\setcopyright{acmcopyright}\acmConference[MM '22]{Proceedings of the 30th ACM International Conference on Multimedia}{October 10--14, 2022}{Lisbon, Portugal}
\acmBooktitle{Proceedings of the 30th ACM International Conference on Multimedia (MM '22), October 10--14, 2022, Lisbon, Portugal}
\acmPrice{15.00}
\acmDOI{XXXXXXX.XXXXXXX}
\acmISBN{978-1-4503-XXXX-X/22/10}

\begin{document}

\fancyhead{}

\title{Cycle-Interactive Generative Adversarial Network for Robust Unsupervised Low-Light Enhancement}


\author{
Zhangkai Ni$^1$, Wenhan Yang$^2$, Hanli Wang$^{1,*}$, Shiqi Wang$^4$, Lin Ma$^3$, Sam Kwong$^{4,*}$
}
\affiliation{%
 % \institution{$^1$Tongji University, $^2$Nanyang Technological University, $^3$Meituan, $^4$City University of Hong Kong}
 $^1$\institution{Department of Computer Science and Technology, Tongji University, Shanghai, China}
 $^2$\institution{School of Electrical and Electronic Engineering, Nanyang Technological University, Singapore}
 % $^3$\institution{Meituan, Beijing, China}
 % $^4$\institution{City University of Hong Kong, Hong Kong}
 \institution{$^3$Meituan, Beijing, China, $^4$Department of Computer Science, City University of Hong Kong, Hong Kong}
 % \city{Hong Kong}
}
% \orcid{0000-0003-3682-6288}
\email{{zkni, hanliwang}@tongji.edu.cn, wenhan.yang@ntu.edu.sg, forest.linma@gmail.com, {shiqwang, cssamk}@cityu.edu.hk}
\thanks{*Corresponding authors: Hanli Wang and Sam Kwong}

\begin{abstract}
Getting rid of the fundamental limitations in fitting to the paired training data, recent unsupervised low-light enhancement methods excel in adjusting illumination and contrast of images.
   However, for unsupervised low light enhancement, the remaining noise suppression issue due to the lacking of supervision of detailed signal largely impedes the wide deployment of these methods in real-world applications.
   Herein, we propose a novel Cycle-Interactive Generative Adversarial Network (CIGAN) for unsupervised low-light image enhancement, which is capable of not only better transferring illumination distributions between low/normal-light images but also manipulating detailed signals between two domains,~\textit{e.g.}, suppressing/synthesizing realistic noise in the cyclic enhancement/degradation process.
   In particular,
  the proposed~\textbf{low-light guided transformation} feed-forwards the features of low-light images from the generator of enhancement GAN (eGAN) into the generator of degradation GAN (dGAN).
   With the learned information of real low-light images, dGAN can synthesize more realistic diverse illumination and contrast in low-light images.
   Moreover, the \textbf{feature randomized perturbation} module in dGAN learns to increase the feature randomness to produce diverse feature distributions, persuading the synthesized low-light images to contain realistic noise.
   Extensive experiments demonstrate both the superiority of the proposed method and the effectiveness of each module in CIGAN.
  
\end{abstract}


%\ccsdesc[300]{Computing methodologies~Computational photography}



% 	\centering
% 	\centerline{\includegraphics[width=1.0\linewidth]{illustration/show.pdf}}
% 	\caption{Examples of our proposed QAGAN model trained on two different unpaired datasets. (\textit{i.e.}, results of models trained using the MIT-Adobe FiveK dataset (red) and the crawled Flickr dataset (green).)}
% 	\label{fig:framework}

\maketitle

\section{Introduction}
\label{sec:intro}
Recent years have witnessed an accelerated growth of capturing devices, enabling the ubiquitous image acquisition in various illuminace conditions.
Typically, images acquired under low-light conditions inevitably degraded by various visual quality impairments, such as undesirable visibility, low contrast, and intensive noise.
Low-light image enhancement aims to restore the latent normal-light image from the observed low-light one to simultaneously obtain desirable visibility, appropriate contrast, and suppressed noise~\cite{wang2019underexposed, yang2020fidelity}.
It greatly improves the quality of images to benefit human vision and can also assist in high-level computer vision tasks, such as image classification~\cite{loh2019getting}, face recognition~\cite{jiang2019enlightengan}, and objection detection~\cite{loh2019getting},~\textit{etc}.
Pioneering low-light image enhancement methods stretch the dynamic range of low-light images, \textit{i.e.} Histogram equalization  (HE)~\cite{abdullah2007dynamic, coltuc2006exact, stark2000adaptive, arici2009histogram}, or adjust the decomposed illumination and reflectance layers adaptively, \textit{i.e.} Retinex-based approaches~\cite{wang2013naturalness, fu2016weighted, jobson1997multiscale, guo2016lime}.

\begin{figure}[t]
	\centering
	\subfigure[Input]{
        \includegraphics[width=4.0cm]{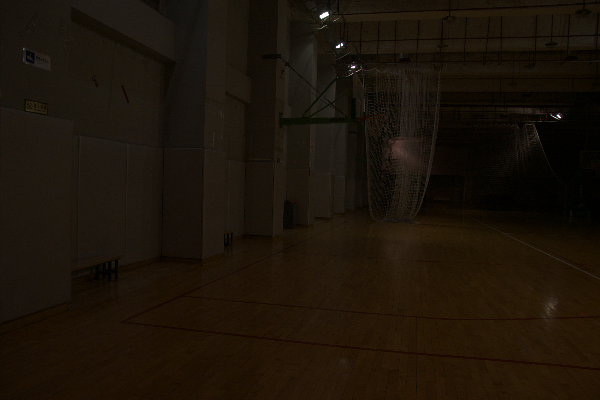}}
	% \hspace{-15pt}
	\subfigure[SICE~\cite{cai2018learning}]{
		\includegraphics[width=4.0cm]{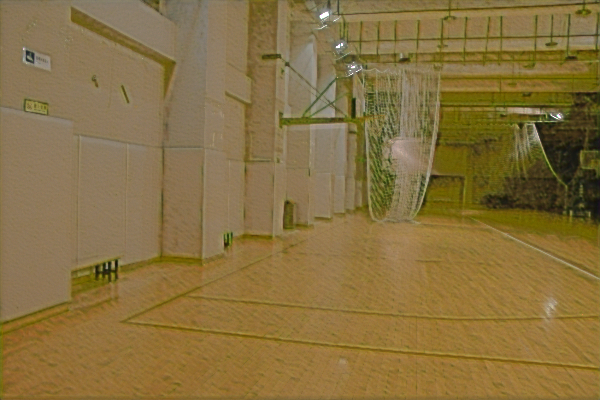}}
	\\  \vspace{-5pt}
	\subfigure[ZeroDCE~\cite{jiang2019enlightengan}]{
        \includegraphics[width=4.0cm]{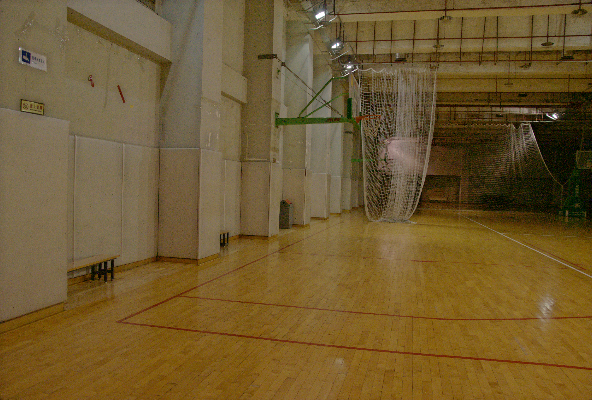}}
% 	\hspace{-5pt}
	\subfigure[CIGAN]{
        \includegraphics[width=4.0cm]{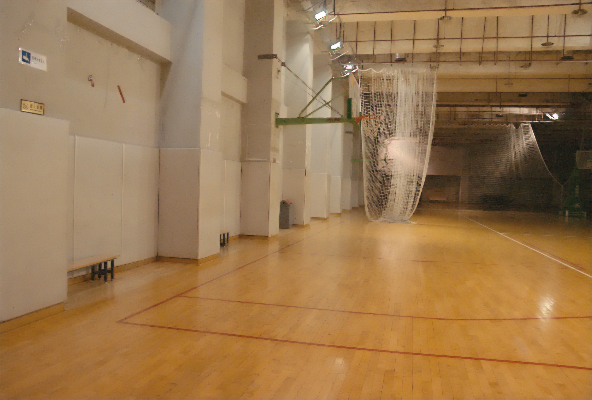}}
	\\
	\vspace{-10pt}
    \caption{
    Visual quality comparison of different methods on a real low-light image in LOL~\cite{wei2018deep}. SICE~\cite{cai2018learning} and ZeroDCE~\cite{guo2020zero} are the leading~\textit{supervised} and~\textit{unsupervised} methods, respectively. Our proposed CIGAN well restores the normal-light image with appropriate illumination and contrast as well as suppressed noise.
	}
	\vspace{-15pt}
	\label{fig:teaser}
\end{figure}

Recently, learning-based approaches have achieved remarkable successes~\cite{cai2018learning, chen2018learning, wang2019underexposed, ni2020towards}. Most of these methods follow the paradigm of supervised learning and heavily rely on the well-prepared paired normal/low-light images to train and evaluate models. However, the commonly seen paired training datasets suffer from their respective limitations. First, \textit{synthesized data} via a simplified simulated imaging pipeline~\cite{lore2017llnet} might fail to capture intrinsic properties of real low-light images. Second, it is quite labor-intensive and time-consuming to create \textit{manual retouching data}~\cite{wang2019underexposed, bychkovsky2011learning} by expert retouchers. It also takes the risk of personal quality bias of the retouchers to adopt such kinds of data as the training data. Third, \textit{real captured data}~\cite{wei2018deep} might capture real degradation but fail to cover diverse scenes and objects in the wild. Besides, the ground truths captured with a pre-defined setting, \textit{i.e.} the exposure time and ISO, might  not be optimal. Therefore, the reliance of supervised methods on the paired data inevitably leads to the domain shift between the training data and testing data in the real world,
further bringing challenges to the generalization on real low-light images.

Recently, a series of unsupervised low-light enhancement methods are proposed. These methods have no reliance on the paired training data and only require two unpaired collections of low/normal-light images. They are built based on the uni-directional generative adversarial network (GAN)~\cite{jiang2019enlightengan} or learnable curve adjustment~\cite{guo2020zero}. These methods achieved promising performances in illumination/contrast adjustment.
%As shown in Fig.~\ref{fig:qualitative_results2},
However, due to the absence of supervision of detailed signal, the quality for some challenging real low-light images with intensive noise are not satisfactory. As a very similar topic, image aesthetic quality enhancement benefits from
CycleGAN~\cite{zhu2017unpaired, chen2018deep, ni2020unpaired} to deliver state-of-the-art performance. We argue that, these CycleGANs neither handle low-light image enhancement problem effectively. First, low-light degradation introduces information lost, which makes the enhancement problem ambiguous. In other words, the mapping between low/normal-light images is~\textit{one-to-many mapping}.
However, CycleGAN can only lead to one-to-one discriminative mapping~\cite{choi2018stargan}.
Second, the intrinsic dimensions of low/normal-light domains are imbalanced as low-light images with intensive noise reflect more complicated properties. The imbalance might disturb the training of CycleGANs, namely that the degradation generator fails to synthesize realistic noise and subsequently the enhancement generator cannot handle the realistic degradation.

In this paper, we propose a novel~\textit{Cycle Interactive GAN} (CIGAN) for unsupervised low-light image enhancement to simultaneously~\textit{adjust illumination, enhance contrast} and~\textit{suppress noise}.
The more comprehensive consideration of image degradation leads to more effective degradation and enhancement processes in cycle modeling. In other words, the more~\textit{realistic and diverse} the low-light images generated in image degradation, the better and more robust the results in image enhancement.
To address the above-mentioned issues of CycleGANs, efforts have been made in three aspects. First, we make the degradation and enhancement generators in our CIGAN~\textit{interact} with each other. More specifically, we propose a novel low-light guided transformation to transfer the features of real low-light images from the enhancement generator to the degradation generator. With the information of different real low-light images as the reference during the whole training process, more diverse low-light images are synthesized, which is beneficial for modeling multiple mappings relationship between low/normal-low images. Second, to handle the domain imbalance issue, we incorporate a novel feature randomized perturbation in the degradation generator. The perturbation applies a learnable randomized affine transform to the intermediate features, which balances the intrinsic dimensions of the features in two domains and is beneficial for synthesizing realistic noise. Last but not least, we design a series of advanced modules to improve the modeling capacities of our CIGAN, such as a dual attention module at the generator side, a multi-scale feature pyramid at the discriminator side, a logarithmic image processing model as the fusion operation of enhancement generator.
Extensive experimental results show that our method is superior to existing~\textit{unsupervised} methods and even the state-of-the-art supervised methods on real low-light images.
To summarize, the main contributions of our paper are three-fold:
\begin{itemize}
% \vspace{-5pt}
\item We propose a novel CIGAN for unsupervised low-light image enhancement to simultaneously~\textit{adjust illumination},~\textit{enhance contrast}, and~\textit{suppress noise}, which excellent in image enhancement and significantly surpasses most previous works in image degradation modeling.
%The more comprehensive consideration of image degradation leads to more effective degradation and enhancement processes in cycle modeling. In other words, the more~\textit{realistic and diverse} the low-light images generated in image degradation, the better and more robust the results in image enhancement.
%That is, \textit{realistic and diverse low-light images} are generated in degradation to help obtain more impressive enhancement results.
\vspace{5pt}
\item We propose a low-light guided transformation (LGT) that allows generators of degradation/enhancement to interact, which helps to generate low-light images with more~\textit{diverse} and~\textit{realistic illumination} and~\textit{contrast}.
%The generators of CIGAN interact via the proposed low-light guided transformation (LGT) for synthesizing more diverse low-light images and benefiting the successive enhancement process.
\vspace{5pt}
\item We propose a learnable feature randomized perturbation (FRP) to produce diverse feature distributions, which makes the generated low-light images with more~\textit{realistic noise} and
 %This process
%obtained more realistic and diverse low-light images will
benefits the low-light image enhancement process.
%We propose a learnable feature randomized perturbation (FRP) to balance the intrinsic feature dimensions of low/normal-light images and synthesize low-light images with more~\textit{realistic noise}.
%\vspace{-5pt}
%\item We design a series of advanced modules to improve the modeling capacities of our CIGAN, such as dual attention module (DAM), multi-scale feature pyramid discriminator (MFPD), a logarithmic image processing (LIP) fusion model.
\end{itemize}

\begin{figure*}[t]
	\centering
	\subfigure{
		\includegraphics[width=1.0\linewidth]{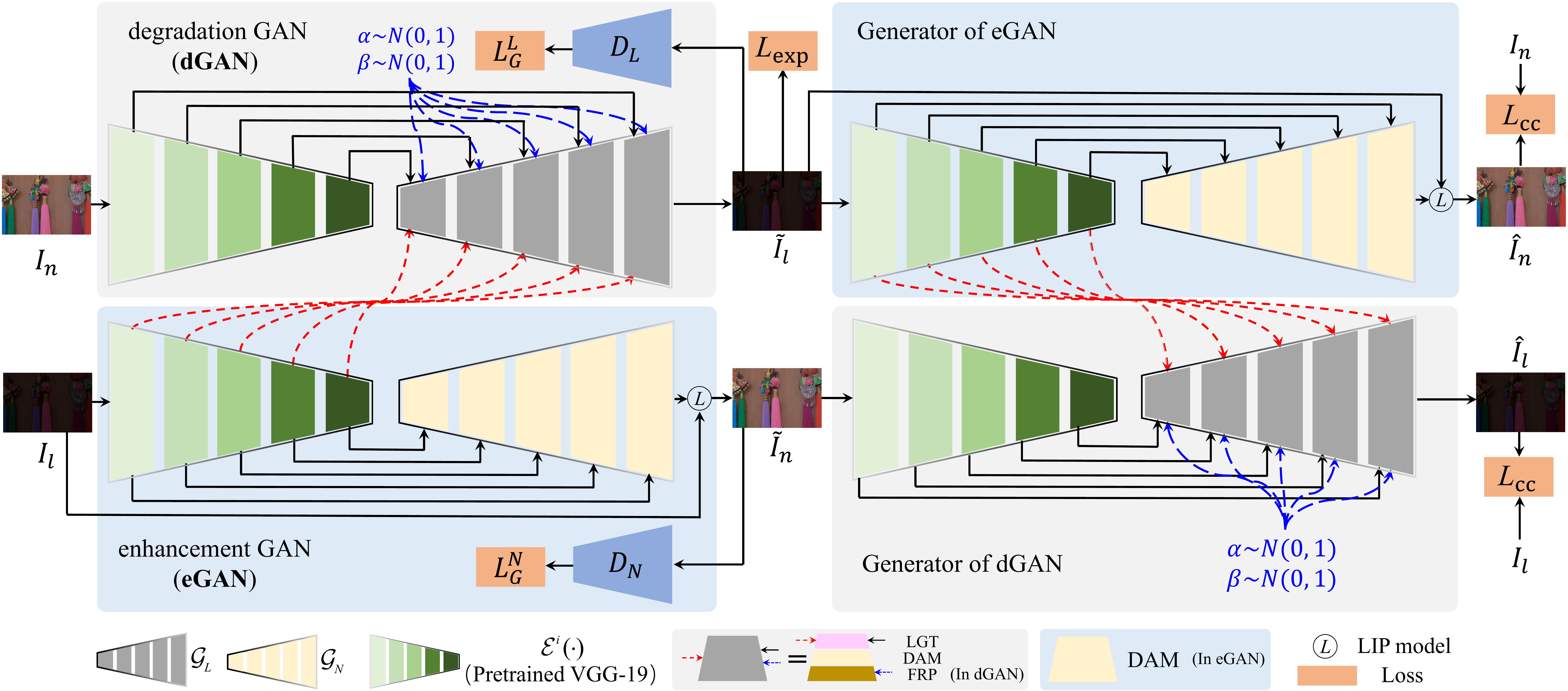}}
	\caption{
	The proposed~\textit{unsupervised} CIGAN consists of complementary dGAN and eGAN.
	(i) The dGAN learns to synthesize realistic low-light images under the supervision of~\textit{unpaired} normal/low-light images. (ii) The eGAN restores normal-light images from synthesized low-light images under the paired supervision generated by dGAN.
	Our CIGAN is different from previous CycleGANs in:
	1) {\color{red}Red dotted line:} the generator of eGAN feed-forwards information of low-light features $\mathcal{E}^i(I_l)$ to that of dGAN via LGT to help dGAN generate more diverse and realistic illumination and contrast;
	2) {\color{blue}Blue dotted line:} Random noise is injected into dGAN via Feature Randomized Perturbation (FRP) to learn to produce more diverse feature distributions for synthesizing more realistic noise.
	The better the synthesized low-light images, the better the eGAN that learns to enhance low-light images towards better illumination, contrast, and suppressed noise.
	%With better synthesized low-light images, eGAN can learn to better enhance low-light images to have better illumination, contrast, and suppressed noise.
	%  The framework of the proposed~\textit{unsupervised} CIGAN consists of two complementary networks: dGAN and eGAN. (i) The dGAN learns to synthesize realistic low-light images under the supervision of~\textit{unpaired} normal/low-light images. (ii) The eGAN, on the contrary, restores normal-light images from synthesized low-light images under the~\textit{paired} supervision generated by dGAN.
    %(i) The dGAN learns to synthesize realistic low-light images under the supervision of~\textit{unpaired} normal/low-light images. To this end, we propose LGT (see Sec.~\ref{sec:lgt}), FRP (see Sec.~\ref{sec:frp}), and $L_{\text{exp}}$ (see Sec.~\ref{sec:objectives}) in dGAN to synthesize low-light images with~\textit{realistic noise} and~\textit{low illumination}. (ii) The eGAN, in contrast, restores normal-light images from synthesized low-light images under the~\textit{paired} supervision generated by dGAN. We propose LIP, DAM (see Sec.~\ref{sec:attention}), and MFPD (see Sec.~\ref{sec:mfpd}) to make eGAN yield high-quality normal-light images.
	}
	\label{fig:framework}
    % \vspace{-15pt}
\end{figure*}

\section{Related Work}
\label{sec:related}

\subsection{Traditional Image Enhancement}
%\vspace{-5pt}

\noindent \textbf{Histogram equalization (HE)}. HE focuses on fitting the illumination histogram to a specific distribution according to local or global statistical characteristics~\cite{abdullah2007dynamic, coltuc2006exact, stark2000adaptive, arici2009histogram}. For example, Arici~\emph{et al.}~\cite{arici2009histogram} cast HE as an optimization problem to improve image contrast while suppressing unnatural effects. Abdullah~\emph{et al.}~\cite{abdullah2007dynamic} proposed a dynamic HE technique using partition operation. Stark~\emph{et al.}~\cite{stark2000adaptive} presented an adaptive contrast enhancement based on generalizations of HE. The main problem of HE is that it easily causes over-enhancement and noise amplification.

\noindent \textbf{Retinex-based approaches}. The Retinex-based method decomposes low-light images into an illumination layer and a reflectance layer to adaptively perform joint illumination adjustment and noise suppression~\cite{wang2013naturalness, fu2016weighted, jobson1997multiscale, guo2016lime}. Wang~\emph{et al.}~\cite{wang2013naturalness} proposed a naturalness Retinex for non-uniform illumination image enhancement. Fu~\emph{et al.}~\cite{fu2016weighted} introduced a weighted variation Retinex that simultaneously estimates the illumination and reflectance layer. These methods have shown satisfying performance in illuminance adjustment, however, hand-crafted constraints are difficult to accurately decompose the low-light image into the illumination and reflection layers, resulting in unnatural visual effects.

%\vspace{-15pt}
\subsection{Learning-based Image Enhancement}
%\vspace{-5pt}

%Deep learning is booming in the field of computer vision.
Low-light image enhancement has achieved great successes with the booming of deep learning~\cite{lore2017llnet, wang2019underexposed, jiang2019enlightengan, yang2020fidelity, guo2020zero}. Broadly speaking, learning-based image enhancement methods can be roughly divided into three categories according to training data: supervised~\cite{cai2018learning, lore2017llnet, ren2019low, wang2019underexposed}, semi-supervised~\cite{yang2020fidelity}, and unsupervised~\cite{jiang2019enlightengan, guo2020zero, ni2020unpaired, xiong2020unsupervised, ni2020towards}. LLNet~\cite{lore2017llnet} is the first attempt to introduce deep learning into the problem of low-light image enhancement. For enhanced performance, various supervised methods are proposed by designing sophisticated network architectures and optimization objects, such as MSR-net~\cite{shen2017msr}, DRD~\cite{wei2018deep}, SICE~\cite{cai2018learning}, DHN~\cite{ren2019low}, and UPE~\cite{wang2019underexposed}. However, these supervised methods are subject to the common restriction of highly dependent on~\emph{paired} data, which limits the performance of these methods on real testing data.
Most recently, Yang~\emph{et al.}~\cite{yang2020fidelity} proposed a semi-supervised low-light enhancement method.
%Most recently, Yang~\emph{et al.}~\cite{yang2020fidelity} proposed a semi-supervised method that jointly uses paired and unpaired data to enhance low-light images.
Jiang~\emph{et al.}~\cite{jiang2019enlightengan} proposed the first unsupervised model based on GAN.
%Jiang~\emph{et al.}~\cite{jiang2019enlightengan} proposed the first unpaired low-light image enhancement model based on GAN.
Guo~\emph{et al.}~\cite{guo2020zero} adopted the no-reference optimization without paired or unpaired data.
%Guo~\emph{et al.}~\cite{guo2020zero} benefited from carefully designed no-reference optimization objects to perform image enhancement without paired or unpaired data.
These unsupervised methods achieve promising performance in illumination adjustment, however, noise suppression has not been considered.
%In this work, our goal is to achieve both~\textit{illumination adjustment} and~\textit{noise suppression} of low-light images in an unsupervised manner.

\begin{figure*}[t]
	\centering
	\subfigure{
		\includegraphics[width=0.96\linewidth]{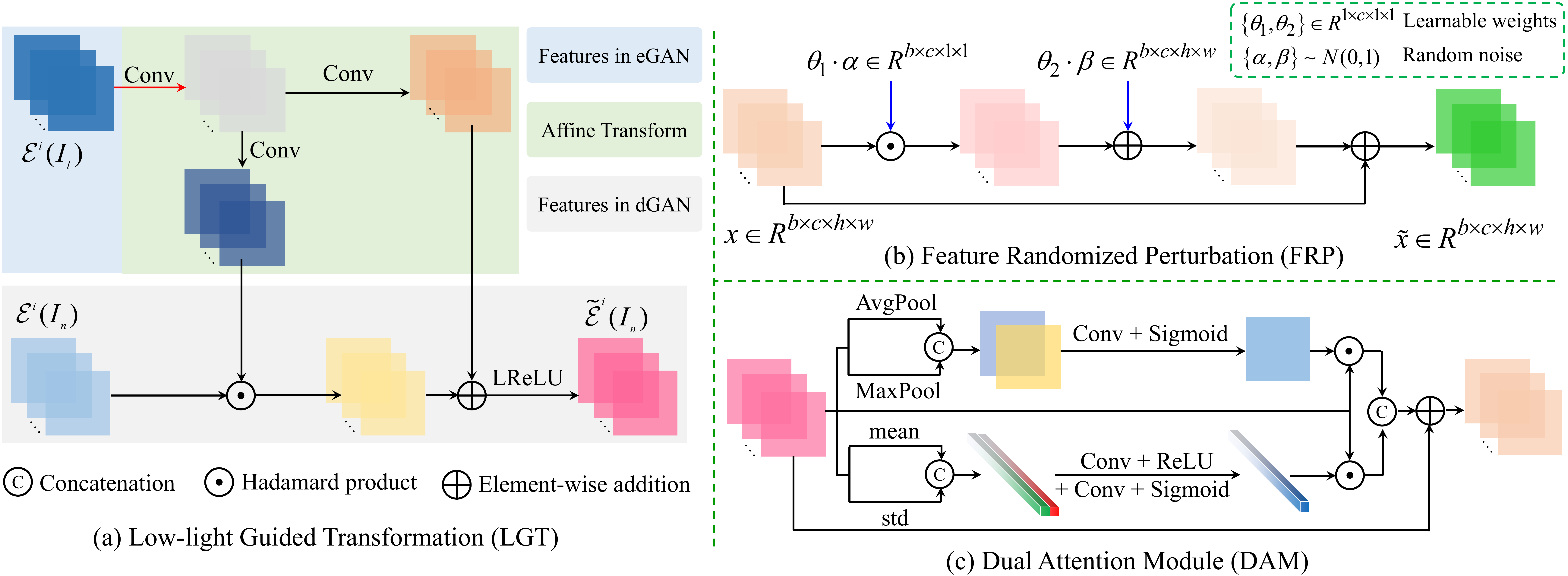}}
	\vspace{-10pt}
    \caption{
    The detailed structure of proposed (a) LGT, (b) FRP, and (c) DAM. The Conv and LReLU are convolution and LeakyReLU operations, respectively.
    %In DAM, the~\textit{AvgPool} and~\textit{MaxPool} are the average-pooling and max-pooling operations along the channel dimension, respectively. Similarly,~\textit{mean} and~\textit{std} are operations for calculating the mean and standard deviation of each channel, respectively.
    The FRP used in dGAN helps synthesize realistic noise. The DAM is used in eGAN and dGAN to effectively model contextual information.
	}
	\label{fig:modules}
    \vspace{-10pt}
\end{figure*}

\section{Method}
\label{sec:methodology}

As shown in Fig.~\ref{fig:framework}, our proposed CIGAN aims to improve the perceptual quality of low-light images by simultaneously adjusting illumination, enhancing contrast and suppressing noise under the supervision of~\textit{unpaired} data.
It consists of complementary degradation GAN (dGAN) and enhancement GAN (eGAN).

\noindent\textbf{1) dGAN}: It aims to synthesize a realistic low-light image $\tilde{I}_l \in \mathbb{L}$ (low-light image domain) from the input normal-light image $I_n \in \mathbb{N}$ (normal-light image domain) with the help of a~\textit{reference} low-light image $I_l \in \mathbb{L}$.
%which is the most challenging part in cyclic modeling for low/normal-light images.
We design two modules to synthesize more realistic low-light images with low-light illumination and contrast as well as intensive noise.
As denoted by the {\color{red}red dotted line} in Fig.~\ref{fig:framework}, \textbf{LGT} (see Sec.~\ref{sec:lgt}\textcolor{red}{-1}) helps dGAN synthesize $\tilde{I}_l$ with the feature information of $I_l$, which preserves the content of $I_n$ while the introduced low-light attributes of $I_l$ makes $\tilde{I}_l$ have more realistic and diverse low-light illumination and contrast.
As denoted by the {\color{blue}blue dotted line} in Fig.~\ref{fig:framework}, \textbf{FRP} (see Sec.~\ref{sec:frp}\textcolor{red}{-2}) learns to inject random noise into features to make the feature distributions more diverse and synthesize more realistic image noise.	
Furthermore, an exposure assessment loss $L_{\text{exp}}$ (see Sec.~\ref{sec:objectives}\textcolor{red}{-1}) is adopted to keep the local average illumination of synthesized low-light images close to a low value.

\noindent\textbf{2) eGAN}: Conversely, it focuses on learning to recover the latent normal-light image $\hat{I}_n$ ($\hat{I}_n \in \mathbb{N}$) from the synthesized low-light image $\tilde{I}_l$.
To make the generator of eGAN yield high-quality normal-light images, we design a flexible logarithmic image processing (LIP) fusion model and a dual attention module (DAM) (see Sec.~\ref{sec:attention}\textcolor{red}{-3}).
%The flow of the proposed algorithm can be boiled down as follows:

%\vspace{-5pt}
\subsection{Model Architecture}
%\vspace{-5pt}
%\noindent \textbf{Generator $\mathcal{G}_L$ of dGAN}.
\noindent \textbf{1) Generator of dGAN}.
%Let $I_N$ and $I_L$ be the input normal-light image and the guided low-light image, respectively. We adopt an encoder $\mathcal{E}$, which is a pretrained VGG-19 network~\cite{simonyan2014very}, to extract multi-scale image feature representations of $I_N$ and $I_L$ as $\mathcal{E}^i(I_N)$ and $\mathcal{E}^i(I_L)$, respectively. The feature representations $\mathcal{E}^i(I_L)$ extracted from $I_L$ are used to adaptively modulate the feature representations $\mathcal{E}^i(I_N)$ obtained from $I_N$, where the modulation is achieved by LGT in generator $\mathcal{G}_L$.
Given an input normal-light image $I_n$ and a reference low-light image $I_l$, we adopt the pre-trained VGG-19 network~\cite{simonyan2014very} $\mathcal{E}(\cdot)$ to extract their multi-scale feature representations as $\mathcal{E}^i(I_n)$ and $\mathcal{E}^i(I_l)$, respectively.
The LGT uses the features $\mathcal{E}^i(I_l)$ extracted from $I_l$ to adaptively modulates the features $\mathcal{E}^i(I_n)$ of $I_n$, which helps to synthesize more diverse illumination and contrast under the guidance of various unpaired reference low-light images.
The DAM is designed to capture context information from spatial and channel dimensions.
The FRP learning randomly perturbs the features of decoder $\mathcal{G}_L$ to help synthesize low-light images with~\textit{realistic noise}.
Therefore, the synthesized low-light image $\tilde{I}_l$ can be expressed as:
% {\setlength\abovedisplayskip{0pt}
% \setlength\belowdisplayskip{0pt}
% \begin{equation}
% \label{equ:dgan}
% \tilde{I}_L = \mathcal{G}_L \Big( \mathcal{E}^i(I_n), \mathcal{E}^i(I_l), \mathcal{T}_i, \mathcal{A}_i, \mathcal{P}_i \Big),
% \end{equation}}
\begin{equation}
\label{equ:dgan}
\tilde{I}_l = \mathcal{G}_L \Big( \mathcal{E}^i(I_n), \mathcal{E}^i(I_l), \mathcal{T}_i, \mathcal{A}_i, \mathcal{P}_i \Big),
\end{equation}where $\mathcal{T}_i$, $\mathcal{A}_i$, and $\mathcal{P}_i$ are LGT, DAM, and FRP at the $i$-th scale, respectively. Basically, the multi-scale features $\mathcal{E}^i(\cdot)$ is relu\textit{i}\_1(\textit{i.e.,} relu1\_1, relu2\_1, relu3\_1, relu4\_1, and relu5\_1, respectively), where the parameters in $\mathcal{E}(\cdot)$ are fixed during the training phase.
%\vspace{1mm}

%\begin{equation}
%\label{equ:dgan}
%\tilde{I}_L = \mathcal{G}_L (\mathcal{E}^i(I_N), \mathcal{E}^i(I_L), \mathcal{T}_\Theta, \mathcal{S}_N),
%\end{equation}
%where $\mathcal{T}_\Theta$ is FRP that encourages the synthesis of low-light images with~\textit{realistic noise}, and $\mathcal{S}_N$ is self-attention module for guided darkening (see Sec.~\ref{sec:attention}). Basically, the multi-scale image feature representations of $\mathcal{E}^i(\cdot)$ are relu\textit{i}\_1(\textit{i.e.,} relu1\_1, relu2\_1, relu3\_1, relu4\_1, and relu5\_1, respectively), where the parameters in $\mathcal{E}$ are fixed during the training phase.

%\noindent \textbf{Generator $\mathcal{G}_N$ of eGAN}.
\noindent \textbf{2) Generator of eGAN}.
%The generator of eGAN works in a complementary way to generator of dGAN, which restores the normal-light image $\hat{I}_n$ from the synthesized low-light image $\tilde{I}_l$:
The generator of eGAN is dedicated to recovering the normal-light image $\hat{I}_n$ from the synthesized low-light image $\tilde{I}_l$:
% {\setlength\abovedisplayskip{0pt}
% \setlength\belowdisplayskip{0pt}
% \begin{equation}
% \label{equ:egan}
% \hat{I}_N = \mathcal{F} \Big( \mathcal{G}_N \big(\mathcal{E}^i(\tilde{I}_l), \mathcal{A}_i\big), \tilde{I}_l \Big).
% \end{equation}}
\begin{equation}
\label{equ:egan}
\hat{I}_n = \mathcal{F} \Big( \mathcal{G}_N \big(\mathcal{E}^i(\tilde{I}_l), \mathcal{A}_i\big), \tilde{I}_l \Big).
\end{equation}where $\mathcal{G}_N$ is the decoder of the generator of eGAN.

Different from most previous methods that subtract the output of the network from the input low-light image to obtain the final enhanced image. We propose a flexible LIP model $\mathcal{F}(\cdot)$ to fuse the input $\tilde{I}_l$ and output $\overline{I}_n$ into one image to combine information from two sources, which are formulated as follows,
% {\setlength\abovedisplayskip{0pt}
% \setlength\belowdisplayskip{0pt}
% \begin{equation}
% \label{equ:lip}
% \hat{I}_N = \frac{\tilde{I}_l + \overline{I}_n}{\lambda + \tilde{I}_l  \overline{I}_n},
% \end{equation}}
\begin{equation}
\label{equ:lip}
\hat{I}_n = \frac{\tilde{I}_l + \overline{I}_n}{\lambda + \tilde{I}_l  \overline{I}_n},
\end{equation}where $\lambda$ is a scalar controlling the enhancement process, which is set to 1 in our work. The proposed LIP model effectively improves the stability and performance of model training (see Sec.~\ref{sec:ablationstudy}).
%Finally, DAM in both dGAN and eGAN is used to model contextual information to further improve performance.

\begin{figure}[t]
	\centering
	\subfigure{
		\includegraphics[width=1.0\linewidth]{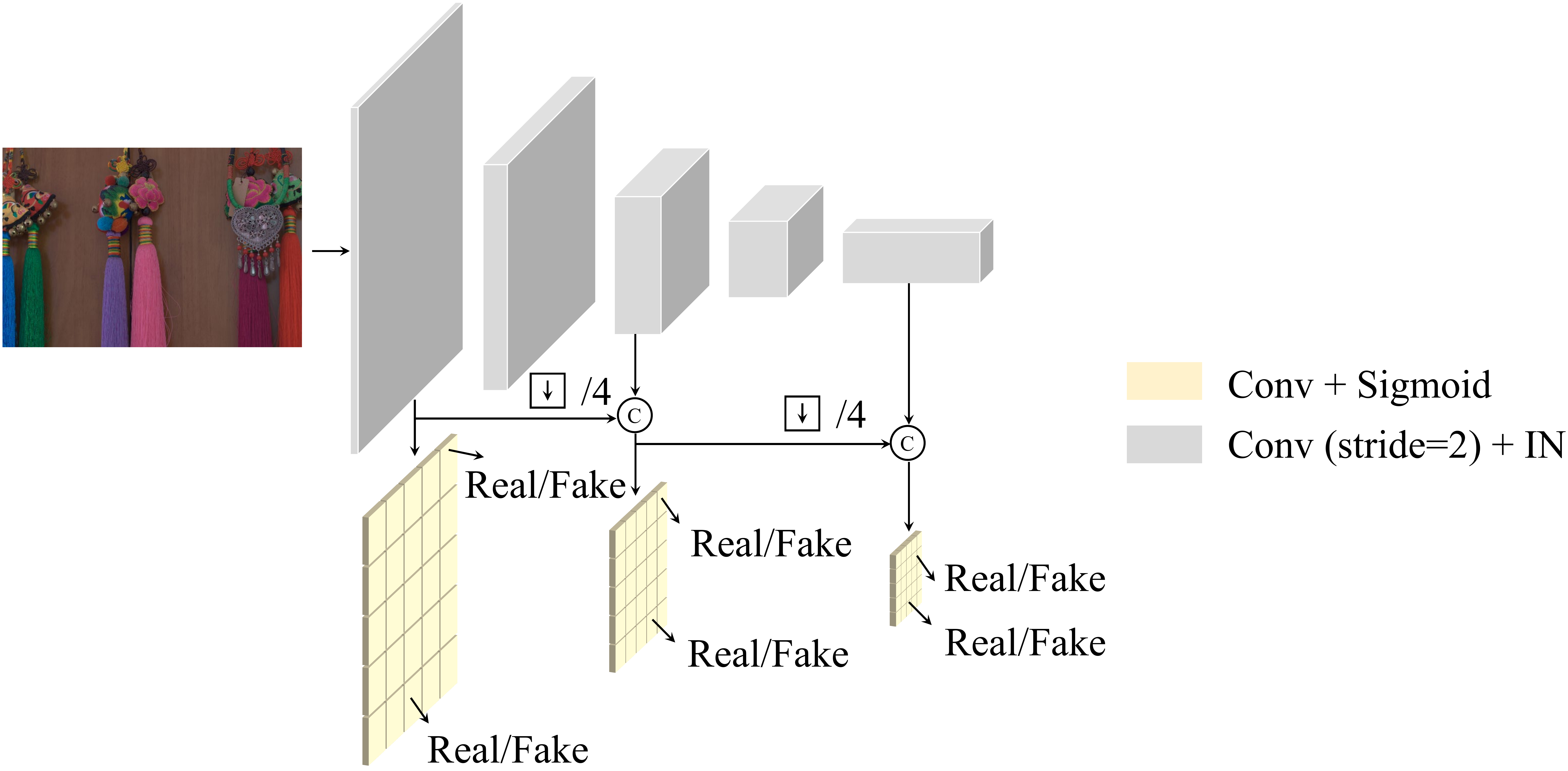}}
	\caption{
    The detailed structure of proposed MFPD.
	}
	\label{fig:mfpd}
    \vspace{-10pt}
\end{figure}

\noindent\textbf{3) Multi-scale Feature Pyramid Discriminator}.
\label{sec:mfpd}
A critical issue associated with GAN is to design a discriminator that can distinguish real/fake images based on local details and global consistency. Our solution is to design a discriminator network that can simultaneously focus on low-level texture and high-level semantic information. Therefore, we propose a multi-scale feature pyramid discriminator (MFPD) as shown in Fig.~\ref{fig:mfpd}. The intermediate layer of the discriminator has a smaller receptive field to make the generator pay more attention to texture and local details, while the last layer has a larger receptive field to encourage the generator to ensure global consistency~\cite{ni2020towards}. In short, the proposed MFPD uses multi-scale intermediate features and a pyramid scheme to guide the generators to generate images with finer local details and appreciable global consistency.

%\vspace{-5pt}
\subsection{Module Design}
\label{sec:moduledesign}
%\vspace{-5pt}

\noindent \textbf{1) Low-light Guided Transformation}. \label{sec:lgt}
%We propose an effective low-light guided affine module (LGT), which is able to adaptively modulate the feature representations of normal-light images to achieve~\textit{fine-grained} low-light image synthesis. The detailed structure of our proposed LGT is shown in Fig.~\ref{fig:modules} (a). To be specific, for the two input feature representations at $i$-th scale: $\mathcal{E}^i(I_N) \in \mathbb{R}^{N\times C\times H\times W}$ from the normal-light image and $\mathcal{E}^i(I_L) \in \mathbb{R}^{N\times C\times H\times W}$ from the low-light image, where $N$ represents the batch size, $C$ is the number of feature map channels, $H$ and $W$ are height and width of feature maps, respectively. The $\mathcal{E}^i(I_L)$ are processed by two convolution layers to generate affine transformation parameters $w(\mathcal{E}^i(I_L))$ and $b(\mathcal{E}^i(I_L))$, where the first convolution layer is shared. The $\mathcal{E}^i(I_N)$ is adaptively modulated by $\mathcal{E}^i(I_L)$ via affine transformation as follows:
We propose a novel low-light guided transformation (LGT) module to transfer the low illumination and contrast attributes of low-light images from the enhancement generator to the degradation generator, which adaptively modulate the features of normal-light images to generate low-light images with more diverse and realistic illumination and contrast. As shown in Fig.~\ref{fig:modules} (a), our LGT has two inputs at the $i$-th scale: the intermediate features $\mathcal{E}^i(I_n) \in \mathbb{R}^{b\times c\times h\times w}$ of the normal-light image and the intermediate features $\mathcal{E}^i(I_l) \in \mathbb{R}^{b\times c\times h\times w}$ from the reference low-light image, where $b$ represents the batch size, $c$ is the number of feature channels, $h$ and $w$ are height and width of feature, respectively. The transformation parameters $w\left(\mathcal{E}^i(I_l)\right)$ and $b\left(\mathcal{E}^i(I_l)\right)$ are learned from the reference features $\mathcal{E}^i(I_l)$ by two convolution layers, where the first convolution is shared. The modulated intermediate features $\tilde{\mathcal{E}}^i(I_n)$ can be produced via affine transformation as follows:
% {\setlength\abovedisplayskip{0pt}
% \setlength\belowdisplayskip{0pt}
% \begin{equation}
% \label{equ:lgt}
% \tilde{\mathcal{E}}^i(I_n) = \mathcal{E}^i(I_n) \odot w\big(\mathcal{E}^i(I_l)\big) + b\big(\mathcal{E}^i(I_l)\big),
% \end{equation}}
\begin{equation}
\label{equ:lgt}
\tilde{\mathcal{E}}^i(I_n) = \mathcal{E}^i(I_n) \odot w\big(\mathcal{E}^i(I_l)\big) + b\big(\mathcal{E}^i(I_l)\big),
\end{equation}where $\odot$ and $+$ are Hadamard element-wise product and element-wise addition, respectively.

Compared with AdaIN~\cite{huang2017arbitrary} using statistical information to perform denormalization on~\textit{channel-wise}, our LGT processes at the~\textit{element level} and provides a flexible way to~\textit{spatially} modulate normal-light image features $\mathcal{E}^i(I_n)$.
%As a consequence, the LGT is able to capture~\textit{fine-grained} visual attributes, which is important for synthesizing realistic low-light images by mimicking the properties of real-world low-light images.
In this way, the proposed LGT incorporates the low illumination and contrast attributes of the reference low-light image into the synthesized low-light image through element-wise affine parameters $w\left(\mathcal{E}^i(I_l)\right)$ and $b\left(\mathcal{E}^i(I_l)\right)$.
%, and can even introduce the noise of the real low-light images.
%\vspace{1mm}

\noindent \textbf{2) Feature Randomized Perturbation}. \label{sec:frp}
%The generator of dGAN aims to synthesize low-light images with low contrast and realistic noise.
The LGT can effectively help to synthesize low-light images with diverse low illumination and contrast but light noise, and achieve relatively promising performance (see Sec.~\ref{sec:ablationstudy}). However, this light noise cannot provide enough information for eGAN to learn to suppress the intensive noise of real low-light images, so that we propose the FRP to make the generator of dGAN synthesize more~\textit{realistic noise}. As shown in Fig.~\ref{fig:modules} (b), the scaling and shifting parameters $\alpha \in \mathbb{R}^{b\times c\times 1\times 1}$ and $\beta \in \mathbb{R}^{b\times 1\times h\times w}$ are sampled from the standard Gaussian distributions, then fused as:
% {\setlength\abovedisplayskip{0pt}
% \setlength\belowdisplayskip{0pt}
% \begin{equation}
% \label{equ:frp}
% \tilde{x} = (1 +  \theta_1 \cdot \alpha) x + \theta_2 \cdot \beta,
% \end{equation}}
\begin{equation}
\label{equ:frp}
\tilde{x} = (1 +  \theta_1 \cdot \alpha) x + \theta_2 \cdot \beta,
\end{equation}where $\{\theta_1, \theta_2\} \in \mathbb{R}^{1\times c\times 1\times 1}$ are two learnable scalar weights, which are learned together with all other parameters of the network by back-propagation. As shown in Fig.~\ref{fig:framework}, we embed the proposed FRP module into generator of dGAN at multiple scales to make the noise of synthesized low-light images close to real low-light images.
%\vspace{1mm}

%\begin{figure}[t]
%	\centering
%	\subfigure{
%	\includegraphics[width=1.0\linewidth]{./illustration/dam.pdf}}
%	\caption{
%    The detailed structure of proposed DAM. The~\textit{AvgPool} and~\textit{MaxPool} are the average-pooling and max-pooling operations along the channel dimension, respectively. Similarly,~\textit{mean} and~\textit{std} are operations for calculating the mean and standard deviation of each channel, respectively.
%	}
%	\label{fig:dam}
%    \vspace{-15pt}
%\end{figure}

\noindent\textbf{3) Attention Module}.
\label{sec:attention}
%We propose a dual attention module (DAM) to model contextual information and feature recalibration.
The DAM is proposed to model contextual information and feature recalibration.
As shown in Fig.~\ref{fig:modules} (c), DAM consists of spatial attention (SA) branch and channel attention (CA) branch. Inspired by SENet~\cite{hu2018squeeze}, both SA and CA perform squeeze and excitation operations in sequence, respectively. Specifically, we use the global average pooling and global max pooling along the channel dimension to compress the feature maps in the SA branch, and adopt mean and standard deviation along the spatial dimension to squeeze the feature maps in the CA branch.

%For images with large spatially varying brightness, generator $\mathcal{G}_L$ is expected to pay more attention to making the brighter regions darker, in contrast, generator $\mathcal{G}_N$ desires to focus on enhancing the darker regions to avoid over-saturation problem in the bright regions. Inspired by~\cite{jiang2019enlightengan}, we design a self-attention module using the luminance component of the YUV color space as:
%{\setlength\abovedisplayskip{5pt}
%\setlength\belowdisplayskip{5pt}
%\begin{equation}
%\label{equ:sa}
%\mathcal{S}_{x} = (1 - Y_x)^\varphi
%\end{equation}}where $\varphi$ is a positive constant used to adjust the curvature of the self-attention map, and $Y_x$ is the luminance component and normalized to [0, 1]. In this work, $\varphi$ is set to 1. The subscript can take $N$ (\textit{i.e.,} $\mathcal{S}_{N}$) or $L$ (\textit{i.e.,} $\mathcal{S}_{L}$) to guide darkening or enhancement. This allows dGAN and eGAN to learn darkening and enhancement based on the brightness of the input image itself, respectively.

\subsection{Training Objectives}
\label{sec:objectives}
%\vspace{-5pt}

\noindent \textbf{1) Exposure Assessment Loss}. We propose the exposure assessment loss to control the exposure consistency between the synthesized low-light images and the real ones. Our insight is to keep the average intensity of local regions of the synthesized low-light images close to a low value.
%Our insight is to keep the intensity of most pixels of the synthesized low-light images close to zero.
Inspired by~\cite{mertens2009exposure}, we formulate $L_\text{exp}$ as:
% {\setlength\abovedisplayskip{0pt}
% \setlength\belowdisplayskip{0pt}
% \begin{equation}
% %\normalsize
% \small
% \label{equ:exploss}
% L_{\text{exp}} = 1 - \text{exp}\big(\frac{-(i-e)^2}{2\sigma^2}\big),
% \end{equation}}
\begin{equation}
%\normalsize
% \small
\label{equ:exploss}
L_{\text{exp}} = 1 - \text{exp}\big(\frac{-(i-e)^2}{2\sigma^2}\big),
\end{equation}where $i$ is the average intensity of a local region, $e$ is the desired intensity, which should be close to a low value, and $\sigma$ controls the smoothness of the Gaussian curve. In our work, $\sigma$, $e$ and the local region size are set to 0.1, 0.1, and $7\times7$, respectively.
%In our work, $\sigma$ equals 0.1, $e$ sets as 0.1,  and the size of the local region is set to $7\times7$.

\noindent \textbf{2) Adversarial Loss}. We adopt the relativistic average Hinge loss GAN (RaHingeGAN) loss~\cite{jolicoeur2018relativistic, ni2020towards} to guide dGAN to synthesize realistic low-light images. The RaHingeGAN loss of dGAN can be formulated as:
% {\setlength\abovedisplayskip{0pt}
% \setlength\belowdisplayskip{0pt}
% \begin{equation}
% %\normalsize
% \small
% % \footnotesize
% \begin{split}
% \label{equ:ladvloss}
% %L^L_{\text{adv}} =
% %\mathbb{E}_{\tilde{I}_L\sim P_{\tilde{L}}}\Big[\text{max}\Big(0, 1 + \big(D_L(\tilde{I}_L) - \mathbb{E}_{I_L\sim P_{L}}D_L(I_L)\big)\Big)\Big] \\
% %+ \mathbb{E}_{I_L\sim P_{L}}\Big[\text{max}\Big(0, 1 - \big(D_L(I_L)-\mathbb{E}_{\tilde{I}_L\sim P_{\tilde{L}}}D_L(\tilde{I}_L)\big)\Big)\Big],  \\
% L^L_{\text{G}} =
% \mathbb{E}_{\tilde{I}_l\sim \mathbb{\tilde{L}}}\Big[\text{max}\Big(0, 1 - \big(D_L(\tilde{I}_l) - \mathbb{E}_{I_l\sim \mathbb{L}}D_L(I_l)\big)\Big)\Big] \\
% + \mathbb{E}_{I_l\sim \mathbb{L}}\Big[\text{max}\Big(0, 1 + \big(D_L(I_l)-\mathbb{E}_{\tilde{I}_l\sim \mathbb{\tilde{L}}}D_L(\tilde{I}_l)\big)\Big)\Big],\\
% L^L_{\text{D}} =
% \mathbb{E}_{\tilde{I}_l\sim \mathbb{\tilde{L}}}\Big[\text{max}\Big(0, 1 + \big(D_L(\tilde{I}_l) - \mathbb{E}_{I_l\sim \mathbb{L}}D_L(I_l)\big)\Big)\Big] \\
% + \mathbb{E}_{I_l\sim \mathbb{L}}\Big[\text{max}\Big(0, 1 - \big(D_L(I_l)-\mathbb{E}_{\tilde{I}_l\sim \mathbb{\tilde{L}}}D_L(\tilde{I}_l)\big)\Big)\Big],
% \end{split}
% \end{equation}}
\begin{equation}
\normalsize
% \small
% \footnotesize
\begin{split}
\label{equ:ladvloss}
L^L_{\text{G}} =
\mathbb{E}_{\tilde{I}_l\sim \mathbb{\tilde{L}}}\Big[\text{max}\Big(0, 1 - \big(D_L(\tilde{I}_l) - \mathbb{E}_{I_l\sim \mathbb{L}}D_L(I_l)\big)\Big)\Big] \\
+ \mathbb{E}_{I_l\sim \mathbb{L}}\Big[\text{max}\Big(0, 1 + \big(D_L(I_l)-\mathbb{E}_{\tilde{I}_l\sim \mathbb{\tilde{L}}}D_L(\tilde{I}_l)\big)\Big)\Big],\\
L^L_{\text{D}} =
\mathbb{E}_{\tilde{I}_l\sim \mathbb{\tilde{L}}}\Big[\text{max}\Big(0, 1 + \big(D_L(\tilde{I}_l) - \mathbb{E}_{I_l\sim \mathbb{L}}D_L(I_l)\big)\Big)\Big] \\
+ \mathbb{E}_{I_l\sim \mathbb{L}}\Big[\text{max}\Big(0, 1 - \big(D_L(I_l)-\mathbb{E}_{\tilde{I}_l\sim \mathbb{\tilde{L}}}D_L(\tilde{I}_l)\big)\Big)\Big],
\end{split}
\end{equation}where $I_l$ is the real low-light image from the domain of low-light images $\mathbb{L}$, and $\tilde{I}_l$ is the synthesized data from the domain of synthesized low-light images $\mathbb{\tilde{L}}$. Similarly, the RaHingeGAN loss of eGAN is expressed as:
% {\setlength\abovedisplayskip{2pt}
% \setlength\belowdisplayskip{2pt}
% \begin{equation}
% %\normalsize
% \small
% % \footnotesize
% %\tiny
% \begin{split}
% \label{equ:eadvloss}
% %L^N_{\text{adv}} =
% %\mathbb{E}_{\tilde{I}_N\sim P_{\tilde{N}}}[\text{max}(0, 1+(D(\tilde{I}_N) - \mathbb{E}_{I_N\sim P_{N}}D_H(I_N)))] \\
% %+ \mathbb{E}_{I_N\sim P_{N}}[\text{max}(0, 1- (D(I_N)-\mathbb{E}_{\tilde{I}_N\sim P_{\tilde{N}}}D_N(\tilde{I}_N)))],
% L^N_{\text{G}} =
% \mathbb{E}_{\tilde{I}_n\sim \mathbb{\tilde{N}}}\Big[\text{max}\Big(0, 1 - \big(D_N(\tilde{I}_n) - \mathbb{E}_{I_n\sim \mathbb{N}}D_N(I_n)\big)\Big)\Big] \\
% + \mathbb{E}_{I_n\sim \mathbb{N}}\Big[\text{max}\Big(0, 1 + \big(D_N(I_n)-\mathbb{E}_{\tilde{I}_n\sim \mathbb{\tilde{N}}}D_N(\tilde{I}_n)\big)\Big)\Big],   \\
% L^N_{\text{D}} =
% \mathbb{E}_{\tilde{I}_n\sim \mathbb{\tilde{N}}}\Big[\text{max}\Big(0, 1 + \big(D_N(\tilde{I}_n) - \mathbb{E}_{I_n\sim \mathbb{N}}D_N(I_n)\big)\Big)\Big] \\
% + \mathbb{E}_{I_n\sim \mathbb{N}}\Big[\text{max}\Big(0, 1 - \big(D_N(I_n)-\mathbb{E}_{\tilde{I}_n\sim \mathbb{\tilde{N}}}D_n(\tilde{I}_n)\big)\Big)\Big],
% \end{split}
% \end{equation}}
\begin{equation}
\normalsize
% \small
% \footnotesize
%\tiny
\begin{split}
\label{equ:eadvloss}
L^N_{\text{G}} =
\mathbb{E}_{\tilde{I}_n\sim \mathbb{\tilde{N}}}\Big[\text{max}\Big(0, 1 - \big(D_N(\tilde{I}_n) - \mathbb{E}_{I_n\sim \mathbb{N}}D_N(I_n)\big)\Big)\Big] \\
+ \mathbb{E}_{I_n\sim \mathbb{N}}\Big[\text{max}\Big(0, 1 + \big(D_N(I_n)-\mathbb{E}_{\tilde{I}_n\sim \mathbb{\tilde{N}}}D_N(\tilde{I}_n)\big)\Big)\Big],   \\
L^N_{\text{D}} =
\mathbb{E}_{\tilde{I}_n\sim \mathbb{\tilde{N}}}\Big[\text{max}\Big(0, 1 + \big(D_N(\tilde{I}_n) - \mathbb{E}_{I_n\sim \mathbb{N}}D_N(I_n)\big)\Big)\Big] \\
+ \mathbb{E}_{I_n\sim \mathbb{N}}\Big[\text{max}\Big(0, 1 - \big(D_N(I_n)-\mathbb{E}_{\tilde{I}_n\sim \mathbb{\tilde{N}}}D_n(\tilde{I}_n)\big)\Big)\Big],
\end{split}
\end{equation}where $\mathbb{N}$ and $\mathbb{\tilde{N}}$ are the real normal-light image domain and synthesized normal-light image domain, respectively.

\begin{table*}[htbp]
    \normalsize
	%\footnotesize
    % \small
    \renewcommand{\arraystretch}{1.1}
    \tabcolsep0.22cm
	\centering
	\caption{Quantitative comparisons of different methods on real low-light test images in \textit{LOL-Real} dataset~\cite{wei2018deep}. EG denotes EnlightenGAN.}
    % \vspace{2pt}
	\begin{tabular}{lccccccccccc}
	\hline
    \hline
	%Metric & BIMEF~\cite{ying2017bio} & BPDHE~\cite{ibrahim2007brightness} & CRM~\cite{ying2017camera}  & DHECE~\cite{nakai2013color} & Dong~\cite{dong2011fast}  & EFF~\cite{ying2017fusion}   & CLAHE~\cite{zuiderveld1994contrast} & LIME~\cite{guo2016lime} & MF~\cite{fu2016fusion} &  CycleGAN~\cite{zhang2020better} &  QAGAN~\cite{ni2020unpaired}\\
    \multirow{2}{*} {Metric} & BIME & BPDHE & CRM  & DHECE & Dong  & EFF   & CLAHE & LIME & MF &  CycleGAN &  QAGAN\\
    {} & \cite{ying2017bio} & \cite{ibrahim2007brightness} & \cite{ying2017camera}  & \cite{nakai2013color} & \cite{dong2011fast}  & \cite{ying2017fusion}   & \cite{zuiderveld1994contrast} & \cite{guo2016lime} & \cite{fu2016fusion} &  \cite{zhang2020better} &  \cite{ni2020unpaired}\\
	\hline
	PSNR     & 17.85   & 13.84   & 19.64   & 14.64   & 17.26   & 17.85   & 13.13  & 15.24   & 18.73  &  18.80  &    18.97   \\
    PSNR-GC     & 24.72   & 19.55   & 24.92   & 16.31   & 20.57   & 24.72   & 16.60  & 17.19   & 20.98  &  23.48  &  24.43    \\
	SSIM     & 0.6526  & 0.4254  & 0.6623  & 0.4450  & 0.5270  & 0.6526  & 0.3709 & 0.4702  & 0.5590 &  0.6316  &  0.6081    \\
	SSIM-GC  & 0.7231  & 0.5936  & 0.6968  & 0.4521  & 0.5715  & 0.7231  & 0.3947 &  0.4905 & 0.5765 &  0.6648  &  0.6513    \\
	\hline
	%Metric &  MR~\cite{jobson1997multiscale} & JED~\cite{ren2018joint} &  RRM~\cite{li2018structure} & SRIE~\cite{fu2016weighted} & DRD~\cite{wei2018deep} & UPE~\cite{wang2019underexposed}  & SICE~\cite{cai2018learning} &   UEGAN  & EG~\cite{jiang2019enlightengan} &   ZeroDCE~\cite{guo2020zero}   &    CIGAN  \\
    \multirow{2}{*} {Metric} &  MR & JED &  RRM & SRIE & DRD & UPE  & SICE &   UEGAN  & EG &   ZeroDCE   &    \multirow{2}{*}{CIGAN}  \\
    {} &  \cite{jobson1997multiscale} & \cite{ren2018joint} &  \cite{li2018structure} & \cite{fu2016weighted} & \cite{wei2018deep} & \cite{wang2019underexposed}  & ~
    \cite{cai2018learning} &   \cite{ni2020towards}  & \cite{jiang2019enlightengan} &   \cite{guo2020zero}   &    {}  \\
	\hline
	PSNR     & 11.67   & 17.33  & 17.34   & 14.45   & 15.48   &     13.27  &  19.40  &   19.60  &  18.23  &  18.07  &    \textbf{19.89}    \\
    PSNR-GC     & 18.47   & 22.87  & 23.18   & 23.91   & 23.87   &     24.57  &  23.63  &  23.65   &  21.99  &  23.64  &    \textbf{26.92}    \\
	SSIM     & 0.4269  & 0.6654 & 0.6859  & 0.5421  & 0.5672  &     0.4521 &  0.6906 &  0.6575    &  0.6165 &  0.6030  &    \textbf{0.7817}    \\
	SSIM-GC  & 0.5158  & 0.7236 &  0.7459 &  0.7075 & 0.7476  &     0.7051 &  0.7250 &  0.6727   & 0.6452  &  0.6739  &    \textbf{0.8189}     \\
	\hline
    \hline
	\end{tabular}
	\label{tab:obj1}
    \vspace{-5pt}
\end{table*}

\noindent \textbf{3) Cycle-Consistency Loss}. It consists of two terms: (1) $L_{\text{con}}$ calculates the $L1$ distance between the input images $I_n$ / $I_l$ and the cycled images $\hat{I}_n$ / $\hat{I}_l$. (2) $L_{\text{per}}$ is formulated as the $L2$ norm between the feature maps of the input images and those of the cycled images, as follows:
% {\setlength\abovedisplayskip{0pt}
% \setlength\belowdisplayskip{0pt}
% \begin{align}
% % \normalsize
% %\small
% %L_{\text{cc}} &= \lambda_{\text{con}} L_{\text{con}} + \lambda_{\text{per}} L_{\text{per}},  \\
% L_{\text{con}} &=  \|I_n - \hat{I}_n\|_1 + \|I_l - \hat{I}_l\|_1,   \nonumber\\
% L_{\text{per}} &=  \|\phi_j(I_n) - \phi_j(\hat{I}_n)\|_2 + \|\phi_j(I_l) - \phi_j(\hat{I}_l)\|_2,
% \end{align}}
\begin{align}
% \normalsize
%\small
%L_{\text{cc}} &= \lambda_{\text{con}} L_{\text{con}} + \lambda_{\text{per}} L_{\text{per}},  \\
L_{\text{con}} &=  \|I_n - \hat{I}_n\|_1 + \|I_l - \hat{I}_l\|_1,   \nonumber\\
L_{\text{per}} &=  \|\phi_j(I_n) - \phi_j(\hat{I}_n)\|_2 + \|\phi_j(I_l) - \phi_j(\hat{I}_l)\|_2,
\end{align}where $\phi_j(\cdot)$ is the feature map of the $j$-th layer of the VGG-19 network~\cite{simonyan2014very}, and relu4\_1 is used in our work.

\noindent \textbf{Total Loss}. The proposed CIGAN is optimized with the following objective,
{\setlength\abovedisplayskip{0pt}
\setlength\belowdisplayskip{0pt}
\begin{equation}
\label{equ:gtotalloss}
L_{\text{G}} = L^L_{\text{G}} + L^N_{\text{G}} + \lambda_{\text{exp}} L_{\text{exp}} + \lambda_{\text{con}} L_{\text{con}} + \lambda_{\text{per}} L_{\text{per}},
\end{equation}}
{\setlength\abovedisplayskip{0pt}
\setlength\belowdisplayskip{0pt}
\begin{equation}
\label{equ:dtotalloss}
L_{\text{D}} = L^L_{\text{D}} + L^N_{\text{D}},
\end{equation}}where $\lambda_{\text{exp}}$, $\lambda_{\text{con}}$, and $\lambda_{\text{per}}$ are positive constants to control the relative importance of $L_{\text{exp}}$, $L_{\text{con}}$, and $L_{\text{per}}$, respectively.

\section{Experiments}
\label{sec:experiments}

In this section, the performance of the proposed method is validated through quantitative and qualitative comparisons as well as user study.
% ~\textit{Moreover, extensive experimental results and analysis of qualitative comparisons and ablation studies are provided in the supplementary material.}

\noindent \textbf{Dataset}. We follow~\cite{yang2020fidelity} to comprehensively evaluate our proposed method on LOL dataset~\cite{wei2018deep} with diverse scenes and much variability. It consists of 689 training image pairs and 100 test image pairs, all of which are captured in real-world scenarios. To meet the requirement of unpaired learning, the training set is divided into two partitions: 344 low-light images and another 345 normal-light images with no intersection with each other. Furthermore, we collect more norm/low-light images from the publicly accessible datasets to expand the training images to 1000 unpaired normal/low-light images.

\noindent \textbf{Baselines}. To carry out an overall comparison and evaluation, the proposed CIGAN is compared with twenty-one classical and state-of-the-art methods, including BIMEF~\cite{ying2017bio}, BPDHE~\cite{ibrahim2007brightness}, CRM~\cite{ying2017camera}, DHECE~\cite{nakai2013color}, Dong~\cite{dong2011fast}, EFF~\cite{ying2017fusion}, CLAHE~\cite{zuiderveld1994contrast}, LIME~\cite{guo2016lime}, MF~\cite{fu2016fusion}, MR~\cite{jobson1997multiscale}, JED~\cite{ren2018joint}, RRM~\cite{li2018structure}, SRIE~\cite{fu2016weighted}, DRD~\cite{wei2018deep}, UPE~\cite{wang2019underexposed}, SICE~\cite{cai2018learning}, CycleGAN~\cite{zhu2017unpaired}, EnlightenGAN~\cite{jiang2019enlightengan}, QAGAN~\cite{ni2020unpaired}, UEGAN~\cite{ni2020towards}, and ZeroDCE~\cite{guo2020zero}, where SICE and UPE, EnlightenGAN and ZeroDCE are the leading~\textit{supervised} and~\textit{unsupervised} methods for low-light image enhancement, respectively.

%\noindent \textbf{Baselines}. To carry out an overall comparison and evaluation, the proposed CIGAN is compared with nineteen classical and state-of-the-art methods, including Bio-Inspired Multi-Exposure Fusion (BIMEF)~\cite{ying2017bio}, Brightness Preserving Dynamic Histogram Equalization (BPDHE)~\cite{ibrahim2007brightness}, Camera Response Model (CRM)~\cite{ying2017camera}, Differential value Histogram Equalization Contrast Enhacement (DHECE)~\cite{nakai2013color}, Dong~\cite{dong2011fast}, Exposure Fusion Framework (EFF)~\cite{ying2017fusion}, Contrast Limited Adaptive Histogram Equalization (CLAHE)~\cite{zuiderveld1994contrast}, Low-Light Image Enhancement via Illumination Map Estimation (LIME)~\cite{guo2016lime}, Multiple Fusion (MF)~\cite{fu2016fusion}, Multiscale Retinex (MR)~\cite{jobson1997multiscale}, Joint Enhancement and Denoising Method (JED)~\cite{ren2018joint}, Refined Retinex Model (RRM)~\cite{li2018structure}, Simultaneous Reflectance and Illumination Estimation (SRIE)~\cite{fu2016weighted}, Deep Retinex Decomposition (DRD)~\cite{wei2018deep}, Deep Underexposed Photo Enhancement (UPE)~\cite{wang2019underexposed}, Single Image Contrast Enhancer (SICE)~\cite{cai2018learning}, CycleGAN~\cite{zhu2017unpaired}, EnlightenGAN~\cite{jiang2019enlightengan}, QAGAN~\cite{ni2020unpaired}, UEGAN~\cite{ni2020towards}, and ZeroDCE~\cite{guo2020zero}, where SICE and UPE, EnlightenGAN and ZeroDCE are the leading~\textit{supervised} and~\textit{unsupervised} methods for low-light image enhancement, respectively.

\noindent \textbf{Evaluation Metrics}. We follow~\cite{cai2018learning, wang2019underexposed, yang2020fidelity} and adopt the most widely-used full-reference image quality assessment (FR-IQA) metrics: PSNR and SSIM~\cite{wang2004image}. And calculated the PNSR and SSIM of the Gamma correction results (\textit{i.e.,} PSNR-GC and SSIM-GC). The PSNR and SSIM quantitatively compare our proposed method with other methods in terms of~\textit{pixel} level and~\textit{structure} level, respectively. The higher the values of the PSNR, SSIM, PNSR-GC, and SSIM-GC, the better the quality of the enhanced images.

\noindent \textbf{Implementation Details}. The network is trained for 100 epoches with a batch size of 10 and images are cropped into $224 \times 224$ patches. The Adam~\cite{kingma2014adam} optimizer with $\beta_1$ = 0 and $\beta_2$ = 0.999 is applied to optimize the network. The learning rate of the generator and the discriminator is initialized to 0.0001, the first 50 epoches are fixed and then linearly decay to zero in the next 50 epochs. The hyper-parameters $L_\text{exp}$, $L_\text{con}$, and $L_\text{per}$ are respectively set to 10, 10, and 1. The spectral norm~\cite{miyato2018spectral} is used to all layers in both generator and discriminator.

\begin{figure*}[htbp]
	\centering
	\subfigure[Input]{
		\includegraphics[width=4.3cm]{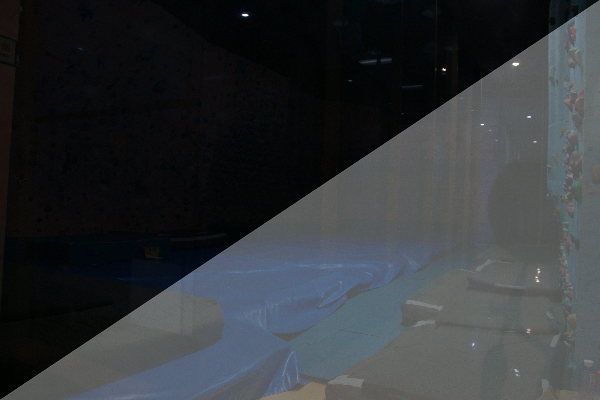}}
	\hspace{-5pt}
	\subfigure[BIMEF~\cite{ying2017bio}]{
		\includegraphics[width=4.3cm]{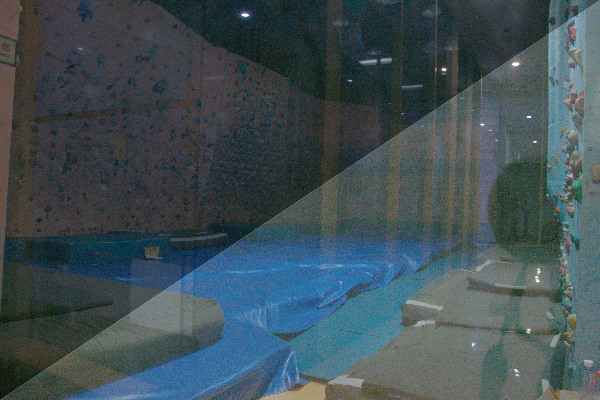}}
	\hspace{-5pt}
	\subfigure[DHECE~\cite{nakai2013color}]{
		\includegraphics[width=4.3cm]{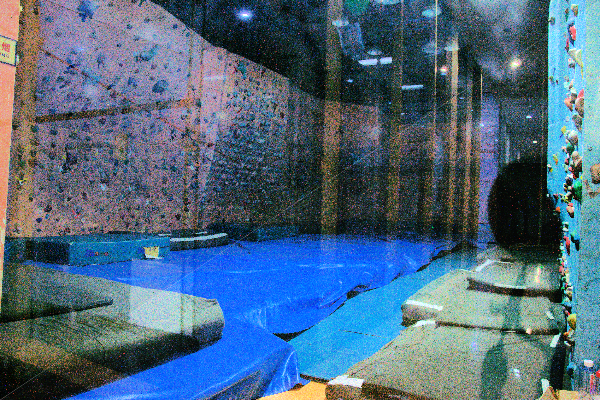}}
	\hspace{-5pt}
	\subfigure[LIME~\cite{guo2016lime}]{
		\includegraphics[width=4.3cm]{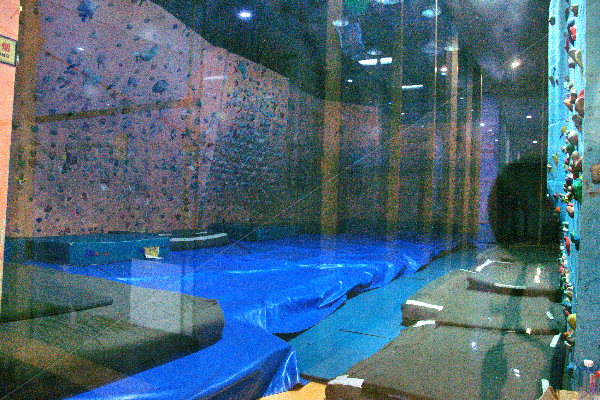}}
	\\ \vspace{-5pt}
	\subfigure[MF~\cite{fu2016fusion}]{
		\includegraphics[width=4.3cm]{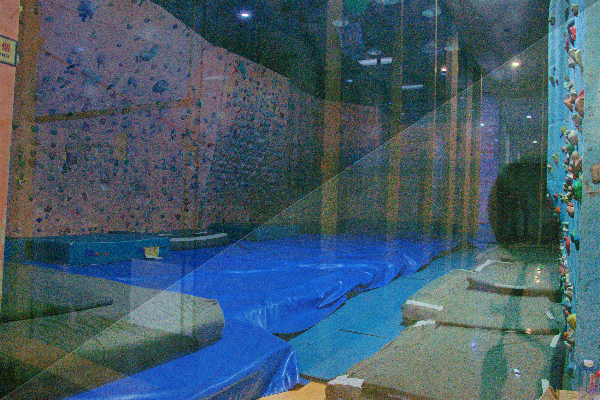}}
	\hspace{-5pt}
	\subfigure[JED~\cite{ren2018joint}]{
		\includegraphics[width=4.3cm]{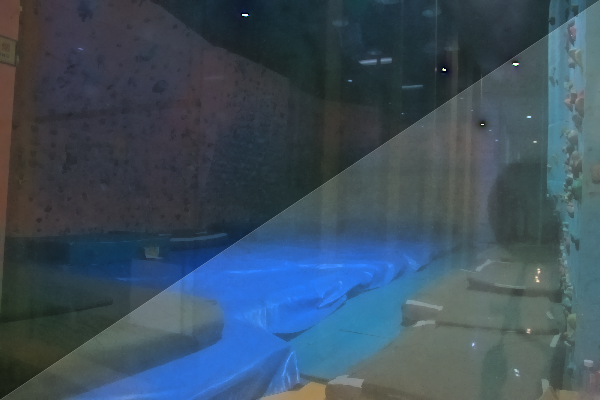}}
	\hspace{-5pt}
	\subfigure[RRM~\cite{li2018structure}]{
		\includegraphics[width=4.3cm]{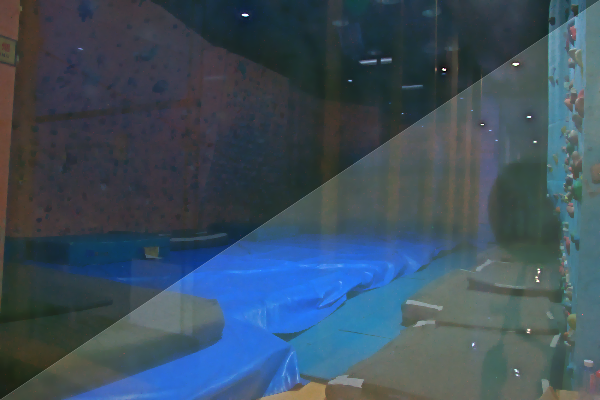}}
	\hspace{-5pt}
	\subfigure[UPE~\cite{wang2019underexposed}]{
		\includegraphics[width=4.3cm]{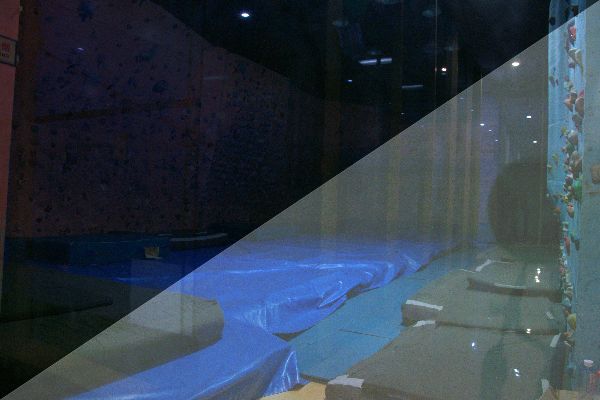}}
	\\ \vspace{-5pt}
	\subfigure[EnlightenGAN~\cite{jiang2019enlightengan}]{
		\includegraphics[width=4.3cm]{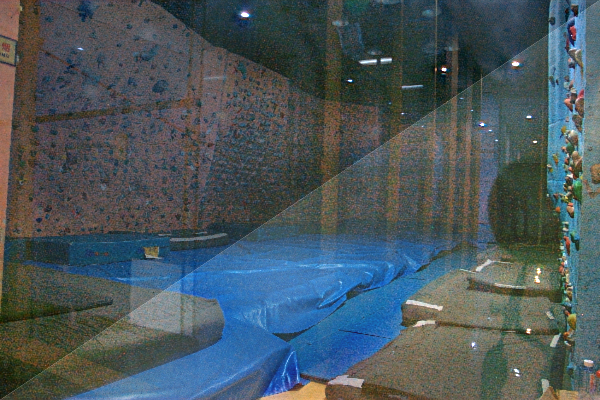}}
	\hspace{-5pt}
	\subfigure[ZeroDCE~\cite{guo2020zero}]{
        \includegraphics[width=4.3cm]{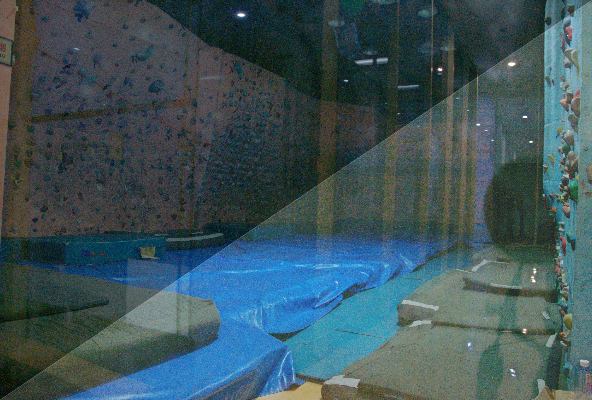}}
	\hspace{-5pt}
	\subfigure[CIGAN]{
		\includegraphics[width=4.3cm]{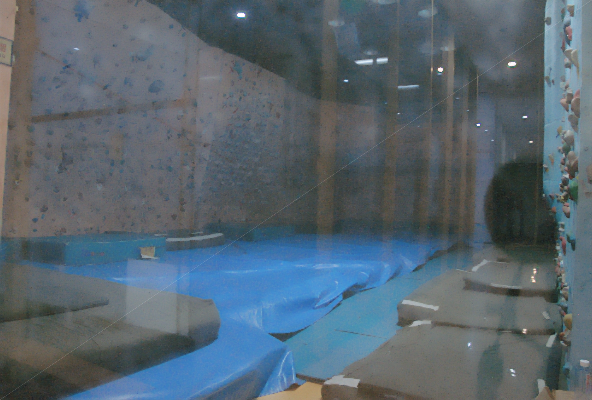}}
	\hspace{-5pt}
	\subfigure[GT]{
		\includegraphics[width=4.3cm]{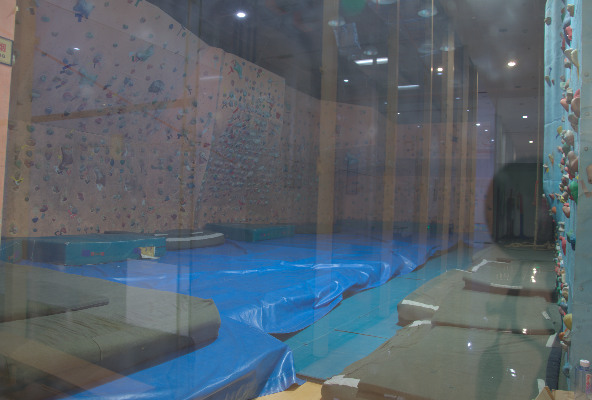}}
	\\ \vspace{-5pt}	
	\caption{Visual quality comparisons of state-of-the-art enhancement methods. Upper left: original results. Lower right: the corresponding results after Gamma transformation correction for better comparison.
	}
% 	\vspace{-15pt}
	\label{fig:qualitative_results1}
\end{figure*}

%\vspace{-5pt}
\subsection{Quantitative Comparison}
\label{sec:quantitative}
% \vspace{5pt}

Table~\ref{tab:obj1} compares the proposed CIGAN with the classical and state-of-the-art methods on LOL dataset~\cite{wei2018deep}. It can be observed that the proposed method outperforms all previous methods in the comparison because it consistently achieves the highest scores in terms of PSNR, PSNR-GC, SSIM, and SSIM-GC. This reveals that the proposed CIGAN is much more effective in illumination enhancement, structure restoration, and noise suppression. From Table~\ref{tab:obj1}, we can see that the proposed method is significantly superior to other state-of-the-art~\textit{unsupervised} methods (\textit{i.e.}, CycleGAN, EnlightenGAN, and ZeroDCE). This is because, on one hand, dGAN makes the attributes of synthesized low-light images consistent with those of real ones, and on the other hand, eGAN is able to restore high-quality normal-light images. Another interesting observation is that the proposed CIGAN even achieves better performance than leading~\textit{supervised} methods (\textit{i.e.}, DRD, UPE, and SICE) trained on a large number of paired images. It is worth noting that the larger PSNR gap between with and without Gamma correction shows that our method can effectively remove intensive noise and restore vivid details.

\subsection{Qualitative Comparison}
\label{sec:qualitative}
\vspace{5pt}

\begin{figure*}[htbp]
	\centering
	\subfigure[Input]{
		\begin{overpic}
        [width=4.3cm] {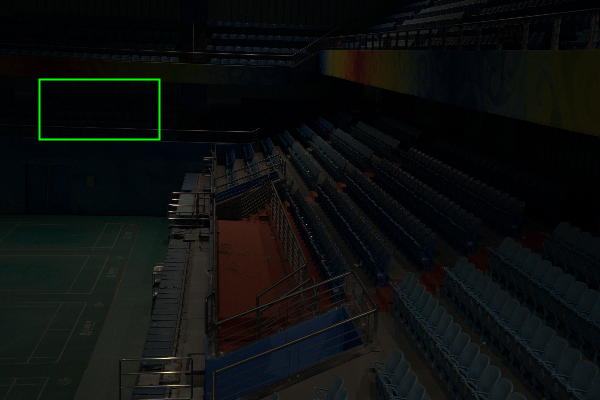}
        \put(0, 0)
        {\includegraphics[scale=0.6]
        {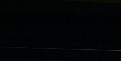}}
        \end{overpic}}
	\hspace{-5pt}
	\subfigure[BIMEF~\cite{ying2017bio}]{
		\begin{overpic}
        [width=4.3cm] {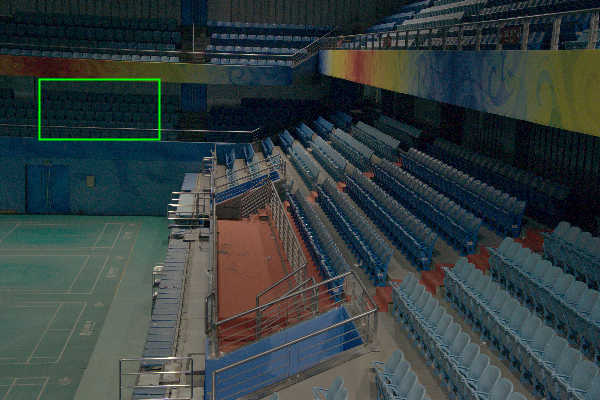}
        \put(0, 0)
        {\includegraphics[scale=0.6]
        {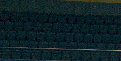}}
        \end{overpic}}
	\hspace{-5pt}
	\subfigure[DHECE~\cite{nakai2013color}]{
        \begin{overpic}
        [width=4.3cm] {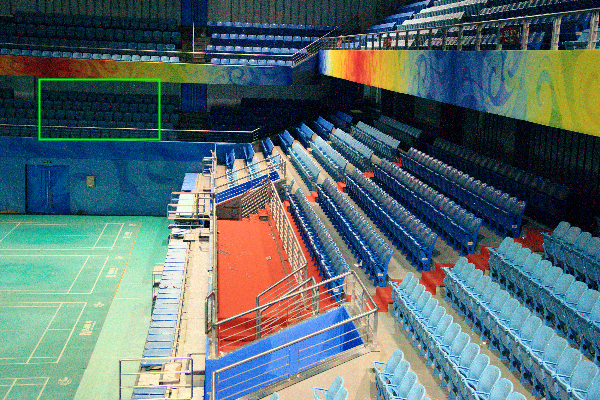}
        \put(0, 0)
        {\includegraphics[scale=0.6]
        {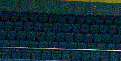}}
        \end{overpic}}
	\hspace{-5pt}
	\subfigure[LIME~\cite{guo2016lime}]{
		\begin{overpic}
        [width=4.3cm] {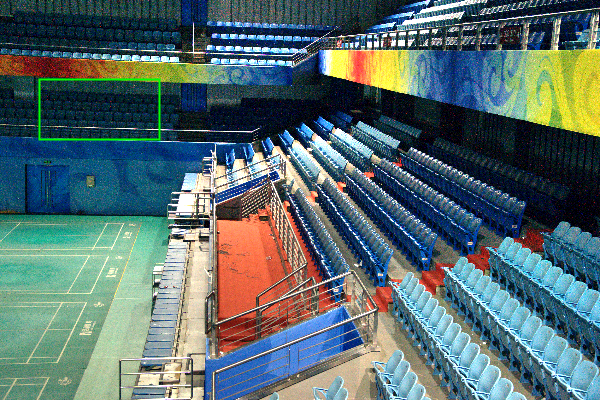}
        \put(0, 0)
        {\includegraphics[scale=0.6]
        {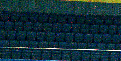}}
        \end{overpic}}
	\\ \vspace{-5pt}
	\subfigure[MF~\cite{fu2016fusion}]{
		\begin{overpic}
        [width=4.3cm] {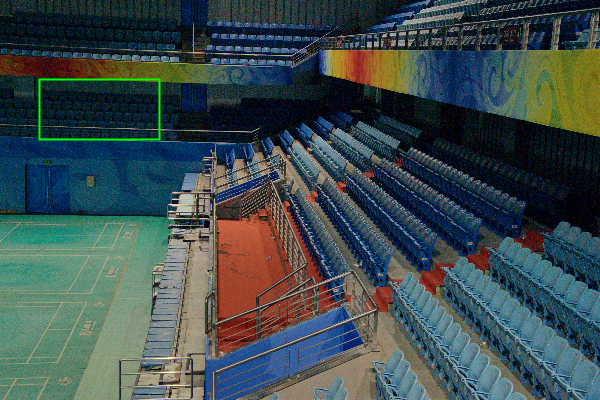}
        \put(0, 0)
        {\includegraphics[scale=0.6]
        {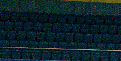}}
        \end{overpic}}
	\hspace{-5pt}
	\subfigure[JED~\cite{ren2018joint}]{
		\begin{overpic}
        [width=4.3cm] {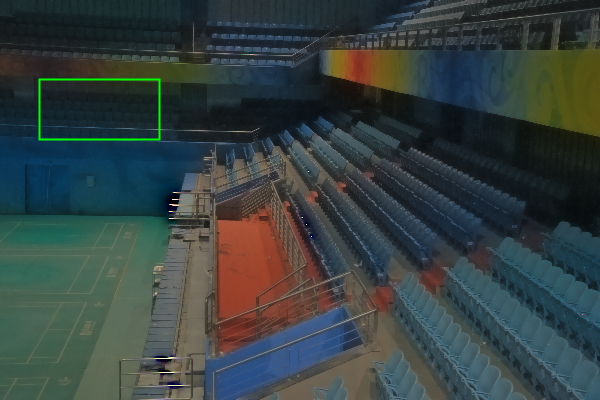}
        \put(0, 0)
        {\includegraphics[scale=0.6]
        {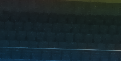}}
        \end{overpic}}
	\hspace{-5pt}
	\subfigure[RRM~\cite{li2018structure}]{
		\begin{overpic}
        [width=4.3cm] {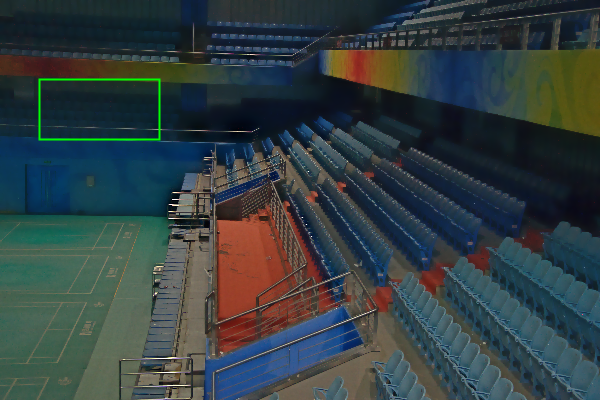}
        \put(0, 0)
        {\includegraphics[scale=0.6]
        {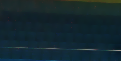}}
        \end{overpic}}
	\hspace{-5pt}
	\subfigure[UPE~\cite{wang2019underexposed}]{
		\begin{overpic}
        [width=4.3cm] {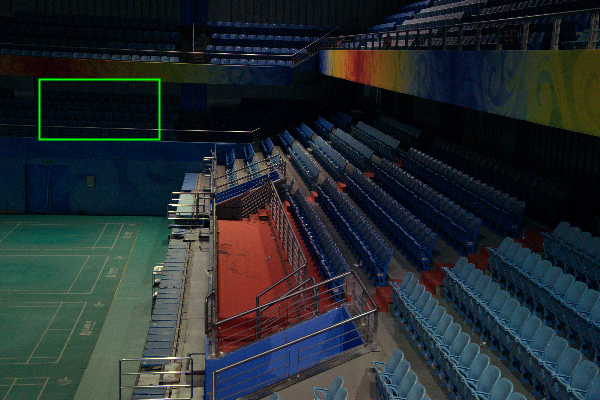}
        \put(0, 0)
        {\includegraphics[scale=0.6]
        {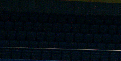}}
        \end{overpic}}
	\\ \vspace{-5pt}
	\subfigure[EnlightenGAN~\cite{jiang2019enlightengan}]{
		\begin{overpic}
        [width=4.3cm] {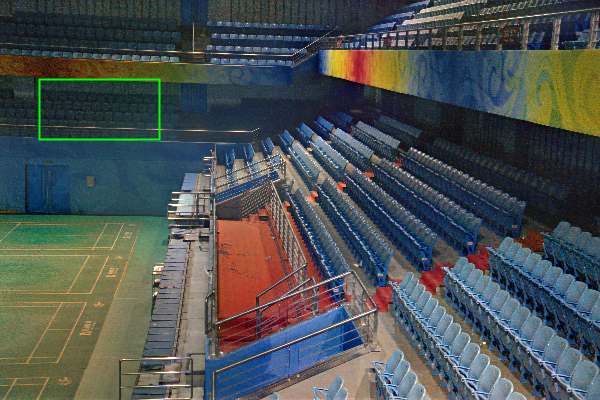}
        \put(0, 0)
        {\includegraphics[scale=0.6]
        {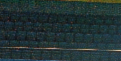}}
        \end{overpic}}
	\hspace{-5pt}
	\subfigure[ZeroDCE~\cite{guo2020zero}]{
		\begin{overpic}
        [width=4.3cm] {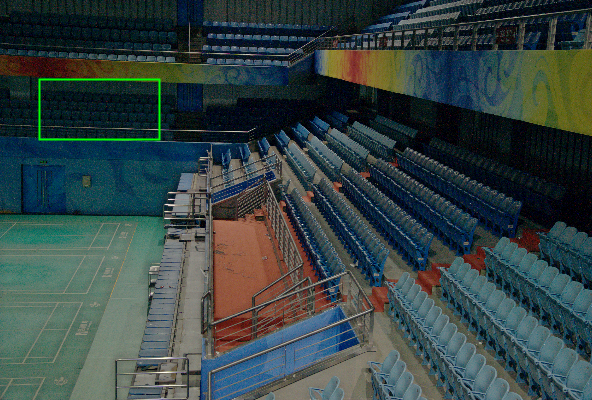}
        \put(0, 0)
        {\includegraphics[scale=0.6]
        {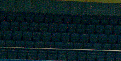}}
        \end{overpic}}
	\hspace{-5pt}
	\subfigure[CIGAN]{
		\begin{overpic}
        [width=4.3cm] {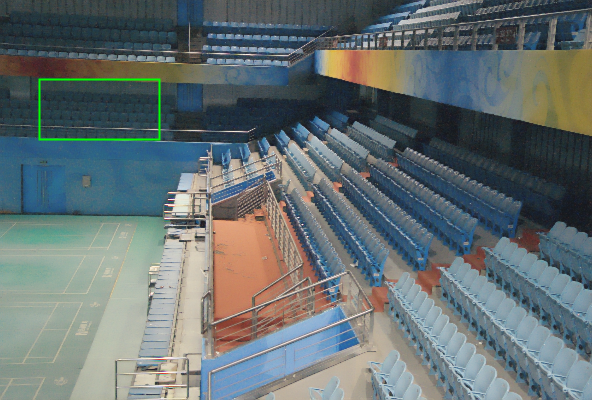}
        \put(0, 0)
        {\includegraphics[scale=0.6]
        {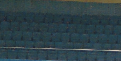}}
        \end{overpic}}
	\hspace{-5pt}
	\subfigure[GT]{
		\begin{overpic}
        [width=4.3cm] {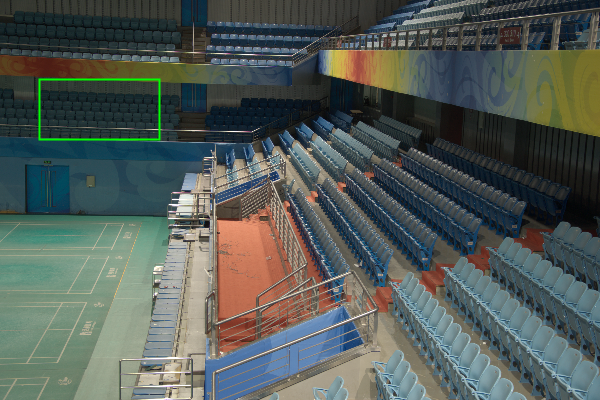}
        \put(0, 0)
        {\includegraphics[scale=0.6]
        {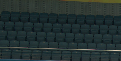}}
        \end{overpic}}
	\\ \vspace{-0pt}	
	\caption{The visual quality comparison for a close-up region of state-of-the-art enhancement methods.
	}
% 	\vspace{-15pt}
	\label{fig:qualitative_results2}
\end{figure*}

Extensive qualitative comparisons have been carried out as shown in Fig.~\ref{fig:qualitative_results1} and \ref{fig:qualitative_results2}. From the enhanced results, we have several insights. First, most of the existing methods (\textit{i.e.}, BIMEF, JED, RRM, UPE, and ZeroDCE) show poor performance in terms of illumination adjustment and detail restoration. The DHECE, LIME, MF, and EnlightenGAN are able to achieve desirable contrast adjustment. However, they also amplify noise and severely degrade visual quality. Second, although several methods especially consider noise suppression (\textit{i.e.}, JED and RRM), they perform unsatisfactory global contrast enhancement and remove many textures and details. In general, our proposed CIGAN achieves favorable visual quality by simultaneously realizing pleasing contrast enhancement and effective noise suppression.

\begin{table}[t]
% 	\footnotesize
    \small
    \renewcommand{\arraystretch}{1.1}
    \tabcolsep0.10cm
	\caption{The results of pairwise comparisons in user study. Each value indicates the number of times the method in the row outperforms the method in the column.}
    % \vspace{5pt}
    \centering
	\begin{tabular}{c|cccccccccc}
		\hline
        \hline
		\multirow{2}{*}{} & DHECE & LIME & UPE & SICE & EG & ZeroDCE  & \multirow{2}{*}{CIGAN} &   \multirow{2}{*}{Total} \\
    &  \cite{nakai2013color} &  \cite{guo2016lime} & \cite{wang2019underexposed}  & \cite{cai2018learning} &  \cite{jiang2019enlightengan} & \cite{guo2020zero}  &   &   \\
        \hline
        DHECE     & -    & 326  & 421  & 511  & 267  & 248  & 16  & 1789   \\
        LIME      & 394  & -    & 452  & 563  & 295  & 269  & 24  & 1997   \\
        UPE       & 299  & 268  &  -   & 391  & 227  & 203  & 19  & 1407   \\
        SICE      & 209  & 157  & 329  & -    & 171  & 143  & 12  & 1021   \\
        EG        & 453  & 425  & 493  & 549  &  -   & 335  & 125 & 2380   \\
        ZeroDCE   & 472  & 451  & 517  & 577  & 385  & -    & 141 & 2543   \\
        CIGAN     & 704  & 696  & 701  & 708  & 595  & 579  & -   & 3983   \\
		\hline
        \hline
	\end{tabular}
    % \vspace{-15pt}
	\label{tab:userstudy}
\end{table}

\begin{figure}[t]
	\centering
    %\vspace{-5pt}
	\subfigure[Input]{
		\includegraphics[width=2cm]{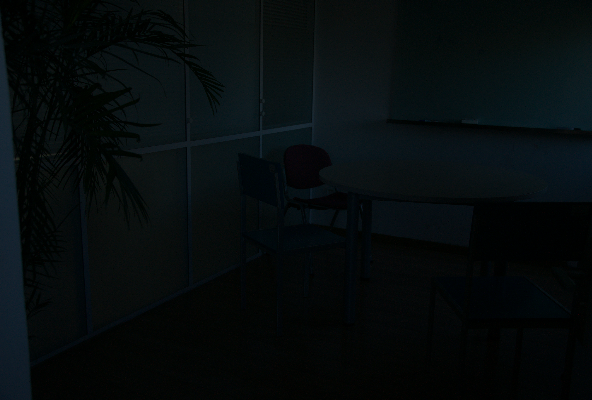}}
	\hspace{-5pt}
	\subfigure[w/o FRP]{
		\includegraphics[width=2cm]{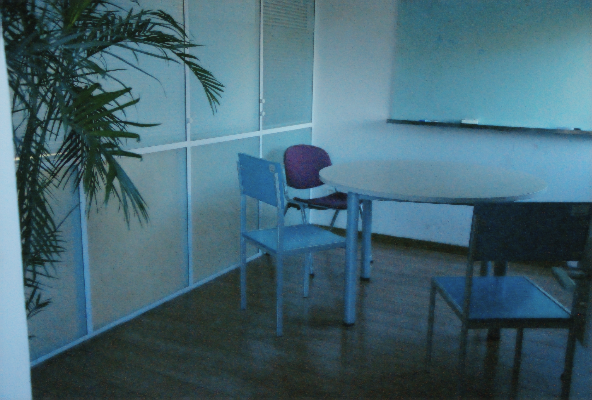}}
    \hspace{-5pt}
	\subfigure[w/o $L_{\text{exp}}$]{
		\includegraphics[width=2cm]{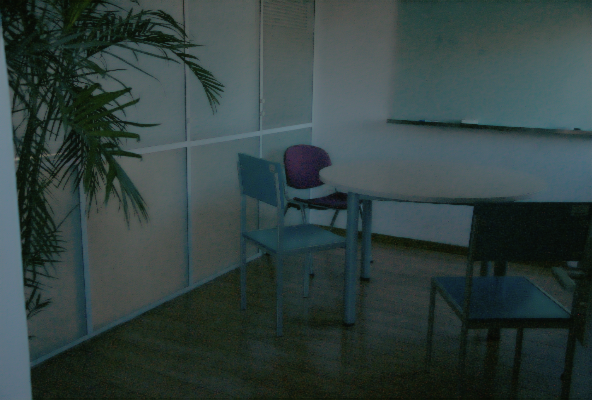}}
	\hspace{-5pt}
	\subfigure[w/o LGT]{
		\includegraphics[width=2cm]{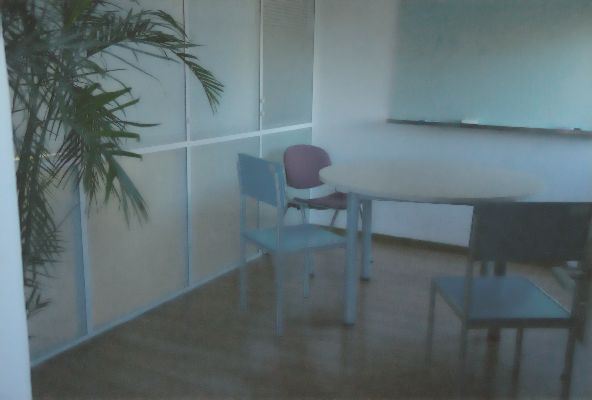}}
	\\ \vspace{-3pt}
    \subfigure[w/o all]{
		\includegraphics[width=2cm]{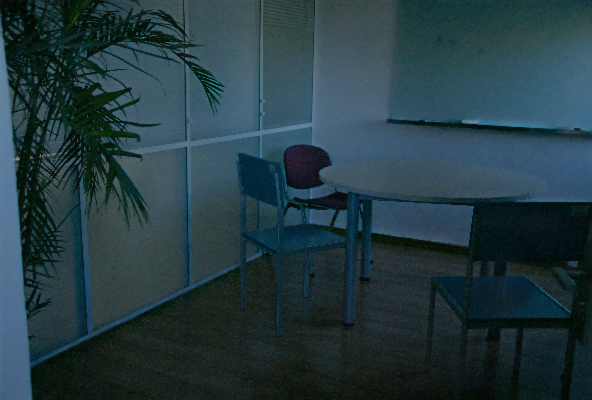}}
	\hspace{-5pt}
    %\subfigure[w/o LGT]{
%		\includegraphics[width=2cm]{illustration/group3/low00703-no-lgat.png}}
%	\hspace{-5pt}
	\subfigure[CIGAN]{
		\includegraphics[width=2cm]{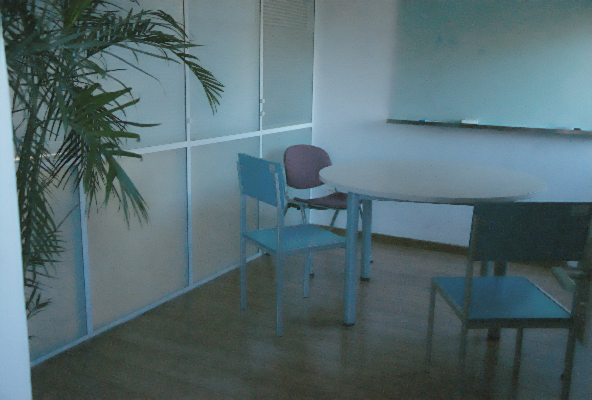}}
	\hspace{-5pt}
	\subfigure[GT]{
		\includegraphics[width=2cm]{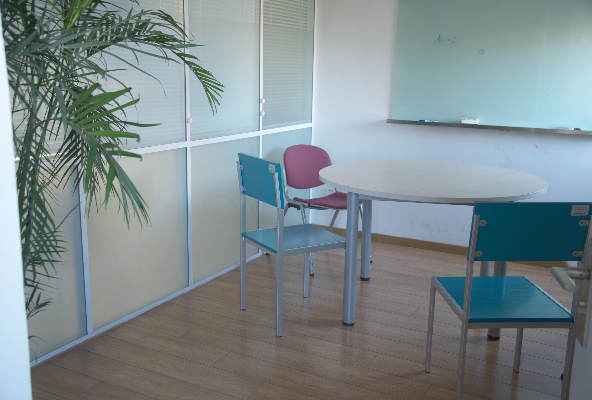}}
	\vspace{-10pt}
	\caption{Ablation study of the effectiveness of three key components (\textit{i.e.}, FRP, LGT, and $L_{\text{exp}}$) in our proposed CIGAN. The (e) w/o all means the CIGAN without FRP, LGT, and $L_{\text{exp}}$ that very similar to vanilla CycleGAN.
	}
% 	\vspace{-15pt}
	\label{fig:ablationstudy}
\end{figure}

\subsection{User Study}
\label{sec:userstudy}
\vspace{5pt}

To study how users prefer the enhanced results of each method, we perform a user study with 24 participants and 30 images of seven methods using pairwise comparisons. Each time the participants are randomly present with the enhanced results of two different methods of the same test image, they are then asked to select their favorite result from the two presented images. Table~\ref{tab:userstudy} tabulates the results of the pairwise comparison, from which we can observe that the enhanced results of the proposed CIGAN are more favorite with users because CIGAN is selected more frequently than the comparison methods. This is consistent with the quantitative and qualitative results, and further consolidates the conclusion that the proposed CIGAN is superior to the state-of-the-art methods.

\begin{table}[t]
\normalsize
%\footnotesize
% \small
\renewcommand{\arraystretch}{1.1}
\caption{Comparison of average PSNR and SSIM performance of different variants of our method on LOL Dataset~\cite{wei2018deep}.}
\vspace{2pt}
\label{tab:ablationstudy}
\tabcolsep0.18cm
\centering
\begin{tabular}{l|cccc}
\hline
\hline
Method  &  PSNR  &    PSNR-GC &   SSIM  &  SSIM-GC \\
\hline
CIGAN w/o LGT  &  18.85   &  25.21   &  0.7488   &  0.7882  \\
CIGAN w/o FRP   &  18.24   &  23.71   &  0.7178   &  0.7573 \\
%CIGAN w/o SAL   &  19.47   &  24.45   &  0.7214   &  0.7546  \\
CIGAN w/o DAM   &  19.25   &  26.28   &  0.7644   &  0.8060  \\
CIGAN w/o MFPD  &  19.55   &  26.36   &  0.7711   &  0.8083  \\
CIGAN w/o LIP   &  19.57   &  24.71   &  0.7641   &  0.7969  \\
CIGAN w/o $L_\text{exp}$ &  19.75	&  24.86 	& 0.7266  & 0.7739  \\
CIGAN  & \textbf{19.89} & \textbf{26.92} & \textbf{0.7817} & \textbf{0.8189}  \\
\hline
\hline
\end{tabular}
% \vspace{-15pt}
\end{table}

\subsection{Ablation Study}
\label{sec:ablationstudy}
\vspace{5pt}

We conduct extensive ablation studies to quantitatively evaluate the effectiveness of each component
%(\textit{i.e.}, LGT, FRP, DAM, MFPD, LIP, and $L_{\text{exp}}$)
in our proposed CIGAN. The variant of CIGAN w/o $L_\text{exp}$ replaces the proposed LIP-based fusion by subtracting the network output from the input low-light image. We perform an ablation analysis on the real low-light image in Fig.~\ref{fig:ablationstudy}. It can be observed that the result produced by CIGAN is obviously better than its variants. Table~\ref{tab:ablationstudy} lists the performance of different variants of our proposed CIGAN on 100 testing images in the LOL dataset in terms of average PSNR, PSNR-GC, SSIM, and SSIM-GC. From Table~\ref{tab:ablationstudy}, we want to emphasize three key components. First, it is critical for LGT to adaptively modulate the normal-light image features with low-light image features. Without it, generator $\mathcal{G}_L$ fails to learn domain-specific properties~\textit{directly} from low-light images, which results in a significant performance degradation. Second, removing the FRP that encourages the synthesis of low-light images with~\textit{realistic noise}, which also leads to a striking performance gap. Last, the proposed exposure assessment loss $L_{\text{exp}}$ plays a key role in synthesizing realistic low-light images that keeps contrast consistent with the real low-light images. All the proposed components lead to better performance, and considering them together enables CIGAN to further improve the quantitative performance towards the best.

%------------------------------------------------------------------------------
\section{Conclusions}
\label{sec:conclusion}
\vspace{5pt}

This paper aims to improve the perceptual quality of real low-light images using~\textit{unpaired} data only in an~\textit{unsupervised} manner. To this end, we propose a novel unsupervised CIGAN, which contains three elaborately designed components: (1) LGT module adaptively modulates normal-light image features with low-light image features to synthesize more diverse low-light images; (2) FRP module encourages the synthesis of low-light images with realistic noise; (3) MFPD improves image quality from coarse-to-fine. Finally, a novel exposure assessment loss is formulated to control the~\textit{exposure} of synthesized low-light images and attention mechanisms are adopted to further improve the image quality. Extensive experiments on real-world low-light images show that our method achieves the superior performance in both quantitative and qualitative evaluations.

%------------------------------------------------------------------------------
\begin{acks}
The authors would like to thank the anonymous referees for their insightful comments and suggestions. 
This work was supported in part by National Natural Science Foundation of China under Grant 61976159, Shanghai Innovation Action Project of Science and Technology under Grant 20511100700.
% This work was supported in part by the Hong Kong RGC General Research Funds under Grant 9042322 (CityU 11200116), Grant 9042489 (CityU 11206317), and Grant 9042816 (CityU 11209819), and in part by the Natural Science Foundation of China under Grant 61672443.

\end{acks}

%------------------------------------------------------------------------------
% \begin{acks}
% The authors would like to thank the anonymous referees for their insightful comments and suggestions. 
% This work was supported in part by the Hong Kong RGC General Research Funds under Grant 9042322 (CityU 11200116), Grant 9042489 (CityU 11206317), and Grant 9042816 (CityU 11209819), and in part by the Natural Science Foundation of China under Grant 61672443.
% \end{acks}

%\begin{acks}
%To Robert, for the bagels and explaining CMYK and color spaces.
%\end{acks}

\bibliographystyle{ACM-Reference-Format}
\balance
\bibliography{CIGAN}

%\appendix

\end{document}